\def\year{2022}\relax
\documentclass[letterpaper]{article} % DO NOT CHANGE THIS
\usepackage{aaai22}  % DO NOT CHANGE THIS
\usepackage{times}  % DO NOT CHANGE THIS
\usepackage{helvet}  % DO NOT CHANGE THIS
\usepackage{courier}  % DO NOT CHANGE THIS
\usepackage[hyphens]{url}  % DO NOT CHANGE THIS
\usepackage{graphicx} % DO NOT CHANGE THIS
\usepackage{natbib}  % DO NOT CHANGE THIS AND DO NOT ADD ANY OPTIONS TO IT
\usepackage{caption} % DO NOT CHANGE THIS AND DO NOT ADD ANY OPTIONS TO IT
\DeclareCaptionStyle{ruled}{labelfont=normalfont,labelsep=colon,strut=off} % DO NOT CHANGE THIS
\usepackage{algorithm}
\usepackage{algorithmic}
\usepackage{booktabs}
\usepackage{multirow}
\usepackage[dvipsnames]{xcolor}
\usepackage{soul}
\usepackage{dirtytalk}
\usepackage{amsmath}
\usepackage{tablefootnote}
\newcommand{\xhdr}[1]{\vspace{1.7mm}\noindent{{\bf #1.}}}

\usepackage{newfloat}
\usepackage{listings}
\floatstyle{ruled}
\newfloat{listing}{tb}{lst}{}
\floatname{listing}{Listing}
\usepackage{lineno}
\title{Othering and Low Status Framing of Immigrant Cuisines in US Restaurant Reviews and Large Language Models}
\title{Othering and Low Status Framing of Immigrant Cuisines in US Restaurant Reviews and Large Language Models}
\author {
    Yiwei Luo,\textsuperscript{\rm 1}
    Kristina Gligorić,\textsuperscript{\rm 2}
    Dan Jurafsky\textsuperscript{\rm 1,2}
}
\affiliations {
    \textsuperscript{\rm 1} Department of Linguistics, Stanford University\\
    \textsuperscript{\rm 2} Department of Computer Science, Stanford University\\
    \texttt{\{yiweil, gligoric, jurafsky\}@stanford.edu}
}
\begin{document}

\maketitle

\begin{abstract}
Identifying implicit attitudes toward food can mitigate social prejudice due to food's salience as a marker of ethnic identity. Stereotypes about food are representational harms that may 
contribute to racialized discourse and %perpetuate prejudice toward ethnic groups and 
negatively impact economic outcomes for restaurants. Understanding the presence of representational harms in online corpora in particular is important, given the increasing use of large language models (LLMs) for text generation and their tendency to reproduce attitudes in their training data.
%Work in sociology has posited two prevailing attitudes toward immigrant food: 1. Immigrant food is objectified as ``exotic'' and ``authentic''%to justify its consumption over American cuisine
%~\cite{johnston2007democracy}, and 2. Within immigrant food, cuisines associated with more assimilated ethnic groups are perceived more positively and given higher status~\cite{ray2007ethnic}. 
Through careful linguistic analyses, we evaluate social theories about attitudes toward immigrant cuisine in a large-scale study of framing differences %between American vs. \textsc{Eur}, \textsc{As}, and \textsc{Lat} cuisines 
in 2.1M English language Yelp reviews. Controlling for factors such as restaurant price and neighborhood racial diversity, we find that immigrant cuisines are more likely to be othered using %objectifying and 
socially constructed frames of authenticity (e.g., \textit{authentic, traditional}), and that non-European cuisines (e.g., Indian, Mexican) in particular are described as more exotic compared to European ones (e.g., French). %while controlling for restaurant price point, neighborhood demographic diversity, and neighborhood income. 
%Furthermore, Asian and Latin American cuisines are framed as more authentic in areas with fewer Asian and Hispanic residents. 
We also find that non-European cuisines are more likely to be described as cheap and dirty, even after controlling for price, and even among the most expensive restaurants. %systematically framed as lower status, being reviewed as less luxurious and in low status terms of cost and hygiene. 
Finally, we show that reviews generated by LLMs reproduce similar framing tendencies, pointing to the downstream retention of these representational harms. Our results corroborate social theories of gastronomic stereotyping, revealing racialized evaluative processes and linguistic strategies through which they manifest.
\end{abstract}

\section{Introduction}

Identifying and understanding implicit attitudes toward food is important for efforts to mitigate social prejudice due to food's pervasive role as a marker of cultural and ethnic identity. Stereotypes about cuisines are a form of representational harm that contribute to racialized public discourse and may in turn perpetuate prejudice toward ethnic groups and negatively impact restaurants' economic outcomes~\cite{luca2016reviews,muchnik2013social}. For example, in the United States, Chinese food is commonly associated with dirty kitchens and cheap takeout, while French food is associated with upscale dining~\cite{hirose2011no,ray2007ethnic}. Understanding the presence of representational harms in online corpora is especially important, given that these corpora are used to train large language models (LLMs) that are increasingly used to generate texts of all kinds.%\footnote{www.cnbc.com/2023/04/25/amazon-reviews-are-being-writ\\ten-by-ai-chatbots.html; %www.theverge.com/2017/8/31/16232\\180/ai-fake-reviews-yelp-amazon; %see also 
%\citet{veselovsky2023artificial}} % and their tendency to reproduce attitudes in their training data.%\footnote{For example, searching for ``exotic food'' on Google Images yields suggestions for Filipino, Thai, and Vietnamese food.}
%stands to reinforce existing prejudices and exacerbate economic inequalities between ethnic groups. 

%Studying attitudes toward food may be especially revealing, since people may be less likely to self-censor when sharing their opinions of food compared to opinions of ethnic groups per se. 

%Work from the sociology of taste has postulated that attitudes toward ethnic foods are intertwined with perceptions of their associated cultures~\cite{ray2007ethnic,ray2017bringing,chhabra2013epitomizing,oleschuk2017foodies}. %, with gastronomical stereotyping both a cause and effect of inaccurate beliefs and narratives about cultural identities and their associated culinary practices. 
%For example, in the United States, Chinese food is commonly associated with cheap take-out and dirty kitchens, while French food is associated with romanticism ~\cite{hirose2011no,lee2012consumer,ray2007ethnic}. 

Prior work on immigrant food attitudes has used coarser-grained signals such as numerical restaurant ratings~\cite{ray2007ethnic} or studied review language within smaller scale datasets%, such as Yelp reviews of restaurants in specific neighborhoods
~\cite{johnston2007democracy,gottlieb2015dirty,gualtieri2022discriminating,hammelman2023paving}. 
%, \textit{Michelin Guide} reviews~\cite{gualtieri2022discriminating}, or writings from food critics~\cite{johnston2007democracy}. 
Meanwhile, larger-scale studies %of attitudes toward immigrant food 
have ignored restaurant and neighborhood-related factors (e.g., price point, neighborhood racial diversity) that may be confounded with review sentiment~\cite{boch2021mainstream,yu2022heightened}. Prior studies have moreover relied on naive word-count-based methods that do not distinguish between contextually different uses, e.g., \textit{a \textbf{stinky} restaurant} vs. \textit{the \textbf{stinky} tofu} (the latter being the name of a Chinese dish).% rather than an evaluative adjective).

We build upon previous work through a study of framing in 2.1M English language Yelp reviews, leveraging NLP parsing techniques to conduct careful linguistic analysis of how customers evaluate restaurants while controlling for restaurant quality and neighborhood demographics (Figure \ref{fig:intro_exs}). Further, since LLMs are increasingly used to generate reviews and texts of all kinds, we investigate the persistence of human stereotypes in LLMs through a controlled review-generation task. 
%since LLMs are increasingly used to generate reviews and texts of all kinds, we investigate whether framing disparities are limited to human reviews, or are in fact reproduced in downstream models trained on online text corpora. % through a prompting study of large language models (LLMs). 
Our research questions are the following: 
\begin{figure}[h!]
    \centering
    \fcolorbox{black}{Aquamarine}{\begin{minipage}{.46\textwidth}
    {\centering \textbf{(Chinese)}} Their menu includes the {\setlength{\fboxsep}{0pt}\colorbox{green}{usual}} American Chinese flairs like sweet and sour (fill in the blank), and (fill in a meat) with black/white or sweet and sour sauce. But they also have a lot of actual {\setlength{\fboxsep}{0pt}\colorbox{lightgray}{authentic}} Chinese dishes [...] %, if you been eating the American Chinese food you can ask the waiter for some help on choosing from this part of the menu. So far in Tampa, they're probably the \hl{best} place to get the roast duck, pork, chicken etc. [...]
    %I usually get their pepper pork(not sure if i remember the name right, but it's pork chops that's been lightly fried and have salt and pepper seasoning on it and is cooked real quickly with hot peppers, onions and green pepper) one of their vegetable dishes(this changes depending on what's in season and if they have it in stock) and another random dish. 
    dim sum is the {\setlength{\fboxsep}{0pt}\colorbox{green}{usual}} stuff, I'm not sure if they push it around in a cart at lunch since I only go there for dinner. The service is pretty %{\setlength{\fboxsep}{0pt}\colorbox{yellow}{good}}
    good, but because I been going there for a while the owner is even %{\setlength{\fboxsep}{0pt}\colorbox{yellow}{nicer}} 
    nicer to my party if she happens to be our server. The place is very %{\setlength{\fboxsep}{0pt}\colorbox{yellow}{nice}} 
    nice and {\setlength{\fboxsep}{0pt}\colorbox{pink}{clean}} [...]
    \end{minipage}}
    \fcolorbox{black}{Apricot}{\begin{minipage}{.46\textwidth}
    {\centering \textbf{(Italian)}} [...] currently my %{\setlength{\fboxsep}{0pt}\colorbox{yellow}{favorite}} 
    favorite local Italian-American restaurant by far. Their food is made with care and focus on presentation. The prices are more than reasonable for the {\setlength{\fboxsep}{0pt}\colorbox{cyan}{breathtaking}} feasts sent to your table with each entree. 
    %{\setlength{\fboxsep}{0pt}\colorbox{yellow}{Wonderful}} 
    Wonderful {\setlength{\fboxsep}{0pt}\colorbox{green}{classic}} dishes and surely the most %{\setlength{\fboxsep}{0pt}\colorbox{yellow}{impressive}} 
    impressive chicken-parm you will find in the state. 
    %There's probably enhancements that could be made to the decor and a few severing tips shared with the waitstaff but I like that Mazzella's doesn't try to be anything it's not.   Just a really %{\setlength{\fboxsep}{0pt}\colorbox{yellow}{satisfying}} 
    %satisfying {\setlength{\fboxsep}{0pt}\colorbox{lightgray}{homey}} restaurant [...]
    \end{minipage}}
    \caption{Example reviews showing frames detected in our analysis (
    %{\setlength{\fboxsep}{0pt}\colorbox{yellow}{yellow}}: positivity; 
    {\setlength{\fboxsep}{0pt}\colorbox{cyan}{blue}}: luxury; {\setlength{\fboxsep}{0pt}\colorbox{green}{green}}: prototypicality; {\setlength{\fboxsep}{0pt}\colorbox{pink}{pink}}: hygiene; {\setlength{\fboxsep}{0pt}\colorbox{lightgray}{gray}}: authenticity). Both customers gave 4 stars %in their reviews 
    and both restaurants are designated as \$ (on the 4-point scale \$ to \$\$\$\$) with the same mean rating.}
    \label{fig:intro_exs}
\end{figure}

\begin{itemize}
\item[\textbf{Q1}] \textit{Framing of immigrant vs. non-immigrant food:} How are immigrant restaurants (i.e. those identified with a cuisine based outside the US)\footnote{We recognize that questions of immigrant identity are deeply nuanced and individual; we thus rely solely on restaurants' self-reported cuisine categories and their associated geographic regions to designate restaurants as immigrant or not. We similarly acknowledge that ``US'' cuisine is itself the result of multiple immigrant influences, though these immigrant origins may not be as salient for contemporary restaurants with US-related labels (e.g., \textit{New American, Southern}) compared to cuisines of regions with high levels of ongoing immigration. Nevertheless, we encourage future work to adopt a more nuanced treatment of individual cuisines.} framed compared to non-immigrant restaurants? %(i.e. those identified with a cuisine based within the US)? 
\item[\textbf{Q2}] \textit{Framing within immigrant foods:} Are cuisines of more assimilated immigrant groups framed differently from those of less assimilated groups? 
\item[\textbf{Q3}] \textit{Framing in synthetic reviews:} Do LLMs transmit the same framing disparities as Yelp reviewers?
\end{itemize}

These questions are informed by work from the sociology of taste. Regarding \textbf{Q1}, the theory of cultural omnivorousness posits that immigrant food is climbing in social status, as part of a broader contemporary shift occurring in other forms of cultural consumption (e.g., music) from a small number of highbrow genres to a more democratic variety~\cite{peterson1997rise}. Crucially, this shift is enabled by the perception and framing of immigrant food as exotic and authentic: by valorizing immigrant food as novel, as well as authentically so, consumers can legitimize their choice of a more omnivorous palate beyond traditional haute cuisine, while still maintaining a pretense of discerning taste over an indiscriminate appetite~\cite{johnston2007democracy}. At the same time, the frames of exoticism and authenticity represent harmful forms of ``othering''~\cite{said1978introduction} that reinforce a mode of outsider cultural consumption that objectifies food in reductive and essentializing ways. 

Regarding \textbf{Q2}, the theory of the ethnic succession of taste argues there are status 
differences in cuisines due to migration patterns and resulting socioeconomic gaps. %between their associated immigrant groups. 
E.g., as Italian immigration to the US slowed in the 20th century and Italian Americans moved up the socioeconomic ladder, so too did the status of Italian food; conversely, Asian and Hispanic food remain lower status, since ongoing Asian and Hispanic immigration continues to populate low-wage jobs, especially in the restaurant industry~\cite{ray2007ethnic,ray2017bringing}. These status differences are reflected similarly to class distinctions in other domains of cultural consumption: whereas highbrow genres tend to receive aesthetic and emotional judgments, lowbrow genres are evaluated on functional and material criteria \citep{bourdieu1987distinction,beagan2015eating,domanski2017omnivorism}. In a culinary context, the latter translates to a concern for hygiene and cost~\cite{zukin2017omnivore,yu2022heightened,williamson2009working,beagan2015eating,hammelman2023paving}. % and cost~\cite{williamson2009working,beagan2015eating}), or both simultaneously~\cite{hammelman2023paving}. 
Together, these theories paint a portrait of US food review discourse that others immigrant cuisines and devalues those of less assimilated ethnicities. %are othered compared to non-immigrant foods, and within immigrant foods, those of less assimilated ethnicities are devalued. %viewed as lower status. 
Finally, we expect LLMs trained on online discourse data to show similar tendencies, in line with work on representational harms, e.g., \citet{crawford2017bias,cheng2023marked}. Thus, our hypotheses are:

\begin{itemize}
\item[\textbf{H1}]Immigrant restaurants are othered (i.e. framed as more exotic \textbf{H1a}, prototypical \textbf{H1b}, and authentic \textbf{H1c}) compared to non-immigrant restaurants.
\item[\textbf{H2}] Restaurants associated with more assimilated immigrant groups are framed in high status terms of luxury \textbf{H2a}; conversely, restaurants of less assimilated groups are framed in low status terms of cost \textbf{H2b} and hygiene \textbf{H2c}.
\item[\textbf{H3}] LLMs reproduce the same framing differences as Yelp reviewers.
\end{itemize}
We test our hypotheses using 2.1M reviews of %immigrant and non-immigrant 
restaurants in 14 US states, controlling for factors such as restaurant price and star rating. We focus on the three largest sources of immigrant food in the US: European cuisine (\textsc{Eur}), Asian cuisine (\textsc{As}), and Latin American cuisine (\textsc{Lat}). % and gastronomic stereotyping.

We find evidence for all 3 sets of hypotheses: immigrant restaurants are overwhelmingly more likely to be described as authentic, though only \textsc{As} and \textsc{Lat} are more likely to be described as exotic. We also find \textsc{As} and \textsc{Lat} are framed as more authentic in zipcodes with fewer Asian and Hispanic residents, respectively, suggesting that othering may be driven by cultural outsiders, rather than the members of an ethnic group themselves.

Further, reviewers are more likely to describe \textsc{As} and \textsc{Lat} as cheap and dirty compared to \textsc{Eur}. These disparities persist even for the most expensive \$\$\$-\$\$\$\$ restaurants (Yelp categorizes restaurants into 4 tiers based on menu prices: \$, \$\$, \$\$\$, \$\$\$\$), supporting the idea of a culinary glass ceiling preventing the cuisines of less assimilated non-European immigrants from attaining the same status as their European counterparts~\cite{ray2017bringing}.

Finally, we find %evidence for \textbf{H3}: 
that GPT-3.5 shows similar framing disparities to human consumers, with immigrant cuisines subject to more othering and non-European immigrant cuisines receiving more low status framing. % compared to \textsc{Eur}.
%we replicate analyses on the reviews of 89 high-volume reviewers (henceforth, ``superyelpers'') across cuisines, finding many of the same framing disparities (see Appendix \ref{ssec:user_controlled}). 

Together, our results corroborate social theories about racialized evaluative processes and
the systematic devaluation of non-white associated cuisines, and demonstrate downstream harms present in LLMs. % and reveal linguistic markers by which their attendant attitudes are reified. 
Our dictionaries and methods\footnote{https://github.com/yiweiluo/immigrant-food-framing/
} can further be applied to other domains of cultural consumption to study perceptions of immigrant cultures in our increasingly globalized society.

%\todo{Discuss why understanding perceptions and framing of food and immigrant culture matters}

%\todo{Discuss how large-scale consumer review data offers a valuable means for understanding such perceptions}

% We computationally evaluate the theory of ethnic succession of taste with a large-scale study of framing differences between US and \textsc{Eur} vs. \textsc{As} and \textsc{Lat} cuisine in Yelp reviews. Specifically, our research questions are the following: 
% \begin{itemize}
%     \item \textbf{Q1:} Are \textsc{As} and Hispanic cuisines framed more negatively compared to \textsc{Eur} and American cuisines?
%     \item \textbf{Q2:} Are \textsc{As} and Hispanic cuisines framed more in terms of stereotypical associations compared to \textsc{Eur} and American cuisines? 
%     \item \textbf{Q3:} Do these framing disparities persist when controlling for restaurant price point, i.e. do we observe a glass ceiling effect?
% \end{itemize}
% We extract dominant per-region frames from 5.2M reviews and find that \todo{summarize main results}, while controlling for neighborhood diversity, neighborhood income, restaurant price, and restaurant average star rating. 

\section{Data and Methods} \label{sec:data_methods}

\begin{figure}
    \centering
    \includegraphics[scale=0.45]{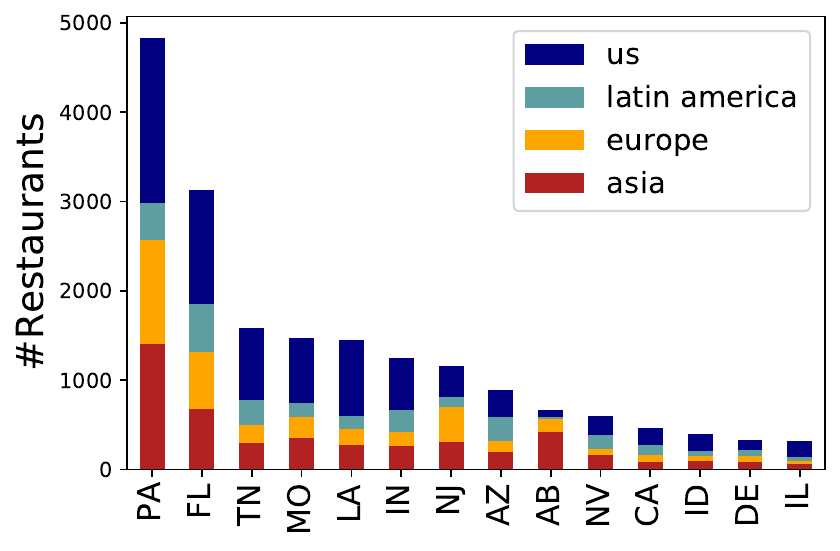}
    \caption{Distribution of restaurants over cuisine, state.}
    \label{fig:per_state_cuisine_dist}
\end{figure}

\xhdr{Real consumer reviews} We use the Yelp open dataset,\footnote{\url{www.yelp.com/dataset}, retrieved January 2023, \copyright Yelp Inc.} containing 5.2M reviews from 2005-2022 of 64K US-based businesses (after subsetting to restaurants only). We additionally exclude chain restaurants, cafes, and fast food. We use businesses' self-declared category tags to obtain cuisine labels for each business (e.g., \textit{Italian}, \textit{Mexican}) and perform analyses on the top 25 %most frequent 
cuisines, excluding \textit{Asian fusion, ethnic food, Caribbean, Middle Eastern,} and \textit{Tex-Mex} as these are difficult to associate with a single region. %, \textit{Indian}). 
%To reduce the number of distinct categories and make comparisons more tractable, 
We then map cuisine labels to broader regions (e.g., \textit{Italian} $\rightarrow$ \textit{Europe}, \textit{Mexican} $\rightarrow$ \textit{Latin America}). Since some restaurants belong to multiple regions (e.g., \textit{Mexican} \& \textit{Spanish}), we subset to restaurants whose cuisines all map to a single region. The above filtering steps yield 2.1M reviews of 16K restaurants based primarily in 14 states, mapping onto the 3 primary sources of immigrant food in the US: European food (\textsc{Eur}), Asian food (\textsc{As}), and Latin American food (\textsc{Lat}), in addition to non-immigrant food (i.e. associated with a US-based cuisine). 
%, \textit{\textsc{As}}). 
%Due to sparsity, we include only the US, Europe, Latin America, and Asia in our analyses. 
Since cuisine distribution within regions is non-uniform, we replicate analyses with the most frequent cuisine per region removed and at the level of individual cuisines (see Appendix). Table \ref{tab:cuisine_stats} summarizes dataset totals and Figure \ref{fig:per_state_cuisine_dist} shows distribution over cuisines per state. We note that Yelp's dataset is highly skewed towards restaurants in Pennsylvania and Florida, and not representative of restaurant concentration in the US overall. 
%(e.g., Italy, Mexico) and that some cuisines do not fall neatly into a single category, so in cases where a restaurant has category tags for multiple regions, %(excluding South America, Africa, and Australia). In cases where a restaurant has category tags belonging to multiple regions (e.g., \textit{\textsc{As}} and \textit{American}) 
%we include the restaurant in both region's samples.

\xhdr{LLM reviews} For synthetically generated reviews, we prompt the \texttt{gpt-3.5-turbo-0613} and \texttt{gpt-3.5-} \texttt{turbo-1106} models via Chat Completion API. %When collecting synthetic reviews, 
For all prompts, we varied price point (\$ -- \$\$\$\$), cuisine (Table \ref{tab:cuisine_stats}), and sentiment. % (very positive, positive, neutral, negative, very negative). 
To collect a variety of review content comparable to Yelp reviews, we also varied the focus of the review for two of our prompts (e.g., entrées, staff, ambiance). %We collected reviews allowing for all potential combinations of the varied dimensions. 
We also %set the \textit{max tokens} parameter to 200 and 
matched the actual distribution of sentiment ratings in Yelp reviews, which skews highly positive (45\% 5-stars). Occasionally, generated reviews contain disclaimer text (e.g., ``As an AI language model, I can say that this customer seems happy with their experience at a French restaurant. They specifically mention that the prices are affordable [...]''). We removed such disclosure text from synthetic reviews so as to analyze only instances of framing from the first person perspective rather than the LLM's meta-commentary (i.e. in the above example, we did not count ``prices are affordable'' toward cost framing). After subsampling to stratify evenly across cuisine regions, our dataset contained 58K total synthetic reviews. See Appendix for further details on our prompting parameters. 

\begin{table}[ht]
\footnotesize
\centering
    \begin{tabular}{lll}
\toprule
\multicolumn{2}{c}{Region} & Cuisine \\
\midrule
\multirow{4}{*}{\rotatebox{90}{Non-imm.}} & US & american traditional (3.6K, 546K), \\ & & american new (3.1K, 561K), cajun/creole\tablefootnote{Yelp collapses Cajun and Creole into a single category. Since both cuisines have immigrant origins (West African, French), we re-do analyses without this category and the main results still hold.} \\ & & (0.5K, 161K), southern (0.5K, 141K), \\ & & soul food (0.3K, 43.7K) \\ \midrule
\multirow{8}{*}{\rotatebox{90}{Immigrant}} & \textsc{Lat} & mexican (1.7K, 184K), latin american \\ & & (0.4K, 42.9K), cuban (0.1K, 14.4K) \\
& \textsc{Eur} & italian (2.2K, 228K), mediterranean (0.5K, \\ & & 63.4K), greek (0.3K, 32.9K), french (0.2K, \\ & & 26.2K), irish (0.1K, 10.1K), spanish (60, 11.6K) \\
& \textsc{As} & chinese (1.6K, 122K), japanese (1.1K, 146K) \\
     & & thai (663, 81.6K), vietnamese (527, 57.3K) \\
     & & indian (442, 46.1K), korean (306, 36.4K) \\
\bottomrule
\end{tabular}
\caption{Summary of Yelp cuisine categories, associated geographic regions, and (\#restaurants, \#reviews) in our dataset.}% We define cuisines associated with \textsc{Lat}, \textsc{Eur}, or \textsc{As} as immigrant and those associated with the US as non-imm.}
\label{tab:cuisine_stats}
\end{table}

\xhdr{Extracting linguistic features} We focus on the framing of three basic restaurant attributes: food (e.g., \textit{chicken, noodles}), staff (e.g., \textit{waiter, server}), and the venue (e.g., \textit{place, atmosphere}). We measure framing via the adjectives %are attributed to or 
modifying tokens from each of the 3 anchor sets. By targeting the framing of specific attributes, we obtain higher precision and more interpretable results compared to aggregate measures of sentiment over the entire text of reviews. For instance, we distinguish true evaluative uses (e.g., \textit{a \textbf{regular} Mexican place}; \textit{the restaurant was \textbf{stinky}}) from false positives (e.g., \textit{I am a \textbf{regular}}, \textit{I had the \textbf{stinky} tofu}). 
We curate anchor sets with WordNet %~\cite{miller1995wordnet} 
and augment the food anchors with a dataset of menu items~\cite{jurafsky2016linguistic}. 
We parse all reviews using spaCy with the Coreferee add-on %~\cite{qi2020stanza} 
and leverage dependency parse relations %such as \texttt{amod}, \texttt{appos}, and \texttt{relmod} 
to retrieve adjectival features, excluding those under the scope of negation (e.g., \textit{\textbf{not} clean}). We then aggregate adjectival features over all anchor sets.\footnote{In early analyses, we examined framing along each attribute individually but found no notable differences.} 
% Examples of dominant per-region and per-cuisine frames are shown in Table \ref{tab:ex_frames}.

%TODO a table with descriptive statistics; fraction reviews that refer to people/food; anchors in parsed output (most frequent ones) per cuisine 

% \begin{table}
%     \begin{tabular}{lllll}
% \toprule
% Region & Cuisine & Frames & & \\
% & & Food & Waitstaff & General \\ 
% \midrule
% US & american(trad.) & hearty, comforting & friendly, attentive & fun, cool, hip \\
%     \end{tabular}
%     \caption{Dominant frames extracted by computing Fightin' Words~\cite{monroe2008fightin} for each region/cuisine compared to all other regions/cuisines.}
% \label{tab:ex_frames}
% \end{table}

%We then identify dominant frames by extracting the adjectives most strongly associated with each cuisine region using the Fightin' Words method~\cite{monroe2008fightin}.
%In addition, we obtain an overall measure of framing valence by computing the proportion of adjectives for each cuisine region %anchored to each set that belonging to the positive emotion lexicon from Linguistic Inquiry and Word Count (LIWC)~\cite{tausczik2010psychological}.
\xhdr{Quantifying framing}\label{ssec:quantifying framing} We quantify framing along broader dimensions of interest with hand-crafted dictionaries. We measure raw counts of linguistic features belonging to each dictionary within review texts to obtain numerical framing scores along each dimension. We next aggregate framing dimensions into broader theoretical constructs as follows: we measure othering as \textbf{exoticism}, which asserts difference and unfamiliarity; \textbf{authenticity}, which asserts faithfulness to something's ethnic origins and 
thus implicitly situates it outside the mainstream~\cite{boch2021mainstream}, and \textbf{prototypicality}, a related form of the ``outsider gaze'' that asserts all X are alike~\cite{rhodes2012cultural,golash2016critical}. 
%More generally, all three forms of framing create socially constructed boundaries that commodify and essentialize food cultures.
We measure %general (de)valuation as \textbf{positive sentiment} (review sentiment is more positive if a reviewer holds something in higher esteem), and finer-grained dimensions of 
perceived high and low status as \textbf{luxury} and fine dining, \textbf{hygiene} (a marker of low perceived status~\cite{zukin2017omnivore,yu2022heightened}), and \textbf{cost/value} (another marker of low perceived status~\cite{williamson2009working}). We acknowledge that these dimensions do not exhaustively measure status/class distinction as a broader construct, though they represent particularly salient dimensions given longstanding theoretical work showing that lowbrow genres are evaluated on functional and material criteria~\cite{bourdieu1987distinction,beagan2015eating,domanski2017omnivorism}.

Our lexicons for exoticism and authenticity are based on social science work~\cite{johnston2007democracy,yu2022heightened,kovacs2014authenticity} and augmented with a thesaurus. For the other framing dimensions, %prototypicality, luxury, cost, and hygiene, 
we use the Empath lexicon induction tool~\cite{fast2016empath}. 
%For positivity, we use the LIWC positive emotion lexicon~\cite{tausczik2010psychological}. 
We manually filter false positives from all lexicons. %remove lemmas that may have distinct meanings in a dining context (e.g., \textit{generous} often refers to portion sizes).
Example lemmas per framing dimension/construct are shown in Table \ref{tab:dict_examples}. See Appendix for full lexicons.

\begin{table}[ht]
\footnotesize
\centering
\begin{tabular}{ccl}
\toprule
    Construct & Frame & Example features \\ \midrule
    \multirow{6}{*}{Othering} & Exoticism & different, distinctive, \\ 
    & & exotic, foreign, odd \\
    & Prototypicality & archetypal, classic, \\ 
    & & stereotypical, usual \\
    & Authenticity & authentic, handmade, \\ 
    & & legit, traditional \\ \midrule
    \multirow{2}{*}{Status (high)} 
    % & Positivity & brilliant, cute, \\
    % & & flawless, romantic \\
    % & Negativity & appalling, bad, boring, \\
    % & & disgusting, hateful, lame \\ 
    & Luxury & alluring, classy,  \\
    & & elegant, posh, refined\\ \midrule
    \multirow{4}{*}{Status (low)} & Cost & affordable, budget, \\
    & & cheap, overpriced 
    \\
    & Hygiene & clean, dirty, grimy, \\ 
    & & nasty, sanitary, stinky
    \\\bottomrule
\end{tabular}
    \caption{Example lemmas for each framing dimension.}
\label{tab:dict_examples}
\end{table}

\begin{figure}[ht]
    \centering
    \includegraphics[scale=0.3]{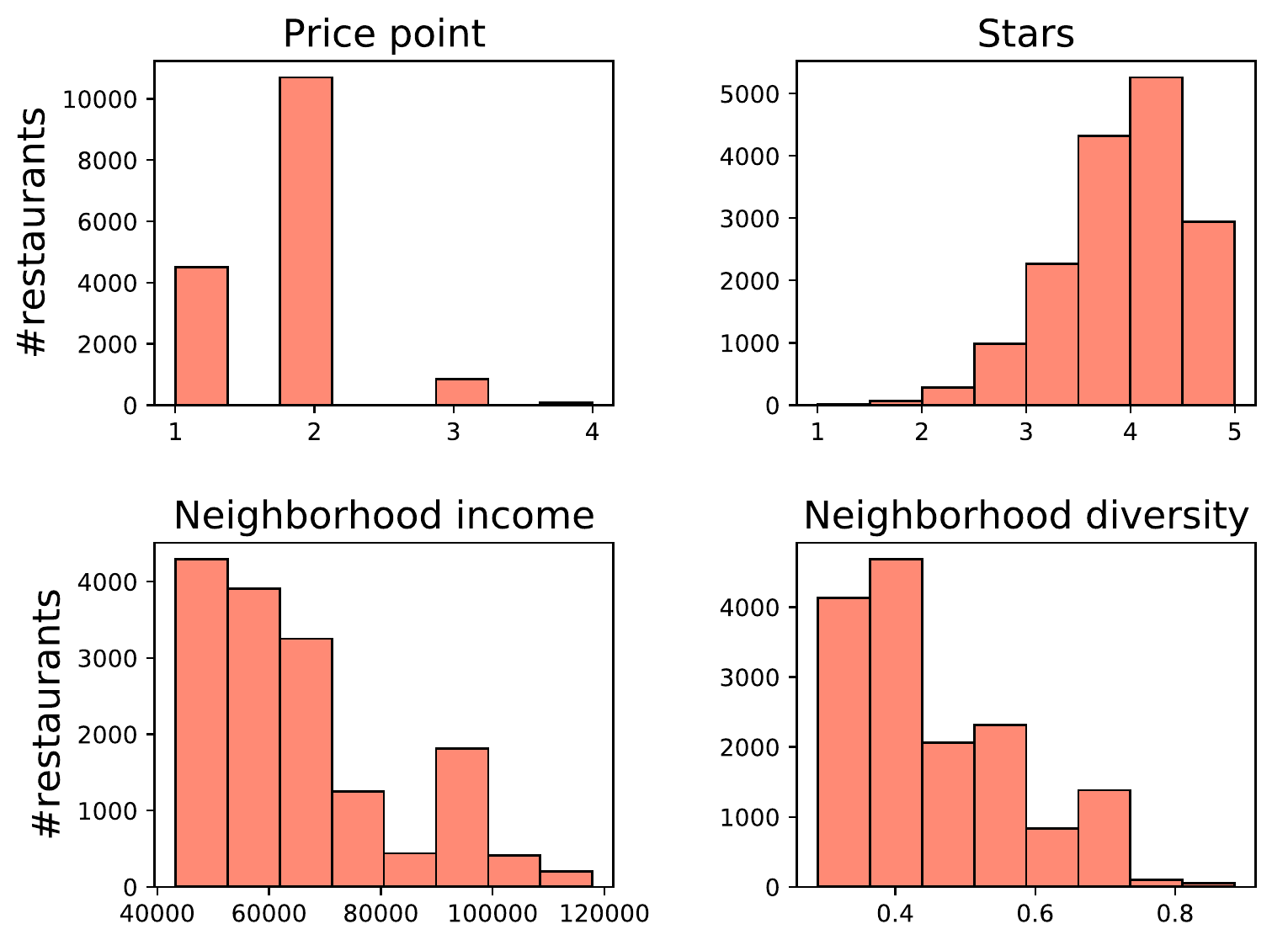}
    \caption{Distribution of restaurants along price point (corresponding to Yelp-designated \$), mean star rating, neighborhood income (2020 USD), and neighborhood racial diversity from 2020 (computed as the Simpson Diversity Index).}
    \label{fig:desc_stats}
\end{figure}

%\subsubsection{Luxury framing}
%correlate our dictionary-based measure with median neighborhood income, expect to see positive correlation
\xhdr{Controlling for confounds in real world reviews} In real world reviews, consumer sentiment is confounded with a number of factors beyond cuisine region, such as review length, %(i.e. the number of words), 
a restaurant's price point and average star rating, and attributes of the restaurant's neighborhood, such as median income and racial diversity. We control for these confounds by including them as covariates in regression analyses to predict framing from cuisine region. We check for multicollinearity among features and find none (variance inflation factors all below 2.0). Figure \ref{fig:desc_stats} shows the distribution of restaurants used in our analyses along restaurant- and neighborhood-attributes. Restaurant price point and mean star rating data is supplied by the Yelp Academic dataset; neighborhood income and racial diversity figures were extracted from 2020 census data after linking each restaurant's geographic coordinates to individual census tracts. Yelp does not release user-specific information, so we were not able to control for user attributes (e.g., age, gender). Instead, we replicate analyses on the reviews of 89 high-volume reviewers and find similar framing disparities across cuisines, within a single user (see Appendix).

% \subsection{Descriptive results}
% \begin{itemize}
%     \item distribution of price level, nb income, nb diversity, mean star rating (confounds we control for)
%     \item proportion of evaluations for food, service, general across cuisines
% \end{itemize}

\xhdr{Qualitatively measuring framing} Separate from our regressions, which as we will see later in the Results section, reveal disparities in aggregate framing scores, we are also interested in qualitative differences in how cuisines are reviewed.
%Beyond measuring aggregate framing scores, we also wish to examine the individual features within each framing dimension most associated with a given cuisine. 
We use the Fightin' Words method \citep{monroe2008fightin}, which measures the strength of association between individual features with a given sample
%computes association strength 
as the weighted log-odds ratio between a feature occurring in a given sample over a reference sample. In our case, we compare a feature's odds of occurring in a review of cuisine $C$ to its odds of occurring in a review of a non-$C$ cuisine. We also include an informative prior in the form of all review texts, and compute the z-score to measure the statistical significance of the association after controlling for variance in feature frequency. Formally, we compute the association strength $\delta$ for a word $w$ with a cuisine $C$ as:

\begin{align} \label{eq:log_odds}
\delta_{w, C}=\cfrac{log\left(\cfrac{L_{w, C}}{L_{w, C'}}\right)}{\sqrt{\cfrac{1}{N_{w, C} + N_{w, P}} + \cfrac{1}{N_{w, C'} + N_{w, P}}}}, 
\end{align}
where $N_{w,C}$ is the count of $w$ in reviews of $C$, $N_{w,C'}$ is the count of $w$ in reviews of non-$C$, and $N_{w,P}$ is the count of $w$ in the prior. $L_{w,C}$ %is the likelihood of $w$ in reviews of $C$, 
and $L_{w,C'}$ %is the likelihood of $w$ in reviews of non-$C$, and 
are defined as follows:
 
\begin{align*} 
L_{w, C}=\cfrac{N_{w, C} + N_{w, P}}{\sum\limits_{x\in C}{} N_{x, C} - N_{w,C} + \sum\limits_{x\in P}{} N_{x,P}-N_{w, P}} \\ 
L_{w, C'}=\cfrac{N_{w, C'} + N_{w, P}}{\sum\limits_{x\in C'}{} N_{x, C'}-N_{w, C'} + \sum\limits_{x\in P}{} N_{x, P}-N_{w, P}}
\end{align*} 
The drawback of Fightin' Words is that we cannot control for confounds like restaurant price as in regressions, but we employ the method to gain insights regarding, e.g., qualitative variation in exoticism framing across cuisines.

\section{Results}

\subsection{Study 1A: Othering of immigrant cuisines in Yelp}

%\todo{TODO: present results with robustness checks: removing over-represented cuisine, over-represented users (positivity measure is not involved in this study)}

\begin{table*}[ht]
\footnotesize
\centering
\begin{tabular}{clll}
\toprule
Frame & Most \textsc{Eur} & Most \textsc{Lat} & Most \textsc{As} \\
\midrule
Exoticism & \textemdash & different (3.9) & exotic (5.0) distinct (2.7) unfamiliar (2.6) \\
Protoypicality & classic (6.4) regular (3.0) & typical (4.4) usual (2.6) & usual (11.7) typical (5.9) standard (5.5) \\ & exemplary (2.0) & & common (4.4) essential (2.2) stereotypical (2.0) \\
Authenticity & homemade (17.5) quaint (5.9) & authentic (51.1) handmade (4.6) & authentic (42.8) traditional (14.6) legit (4.6) \\ 
& true (2.6) rustic (2.1) & & unassuming (3.6) modest (2.1) \\ 
\bottomrule
\end{tabular}
\caption{Features within each othering frame most associated with each immigrant cuisine. Association strengths measured as z-scores of the weighted log odds ratio between a feature occurring in a review of a cuisine over all other cuisines (see Eq. \ref{eq:log_odds}).}
\label{tab:othering_frames_lor}
\end{table*}

\begin{figure}[ht]
\centering
\includegraphics[scale=0.32]{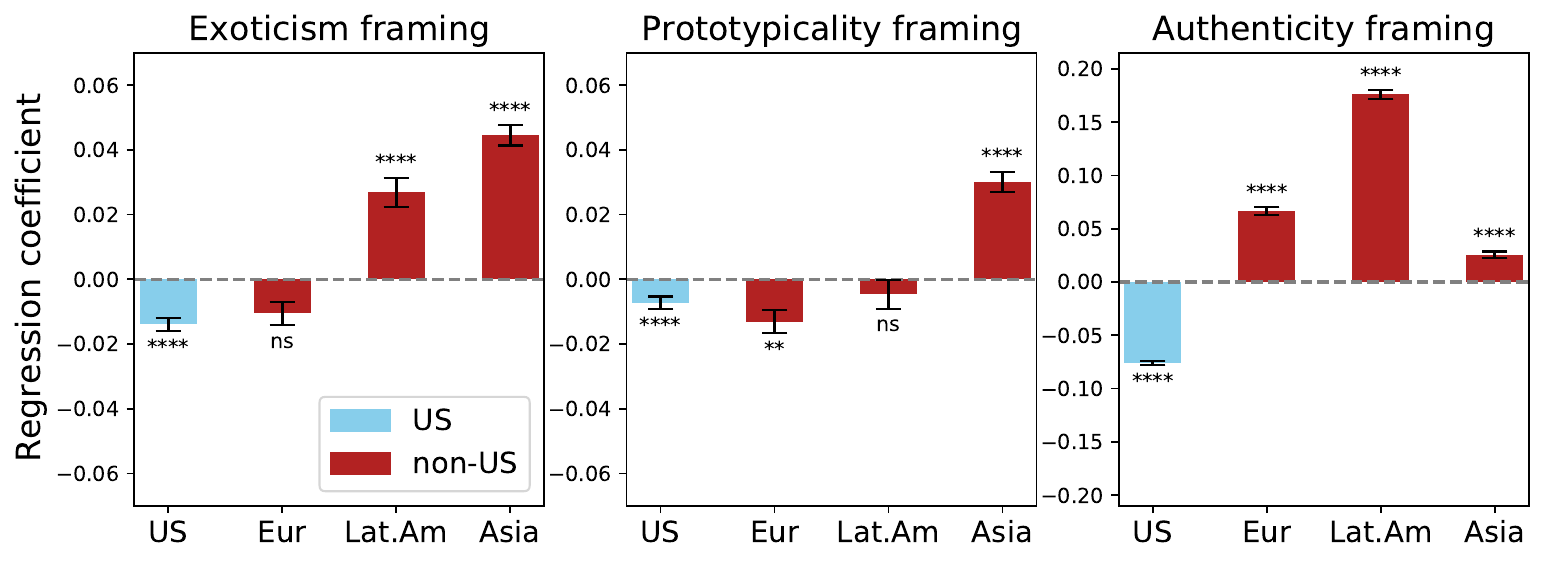}
    \caption{\textbf{Othering of immigrant cuisine}: Linear regression coefficients predicting othering in review text from cuisine region, showing immigrant cuisines receive more othering.  %($\beta$\textsubscript{non-US exot.} (the effect of being a non-US restaurant on exoticism framing)=0.032, p=0.0; $\beta$\textsubscript{non-US prot.}=0.015, p=0.0; $\beta$\textsubscript{non-US auth.}=0.16, p=0.0). %Regressions control for review length, restaurant mean star rating, restaurant price point (reference level: \$\$), neighborhood median income and racial diversity. 
    Note the different y-scale for Authenticity framing. Error bars are 95\% CIs. Significance values for \textsc{US} (the reference level) indicates whether being \textsc{US} has a significant effect on the outcome variable, other values indicate whether a cuisine is significantly different from \textsc{US} with respect to its effect on the outcome. ns=p$>$0.05, **=p$<$0.01, ***=p$<$0.001.}
    \label{fig:study1_results}
\end{figure}

To test our first set of hypotheses concerning othering of immigrant restaurants, namely that immigrant restaurants are framed as more exotic \textbf{H1a}, prototypical \textbf{H1b}, and authentic \textbf{H1c} compared to non-immigrant restaurants, we fit separate linear regression models 
%\footnote{Regressions are fit 
using the following equation: \begin{equation}\label{eq:regression}
    Y = \beta_{0}\cdot C + \beta_{1}\cdot l + \beta_{2}\cdot p + \beta_{2}\cdot s + \beta_{3}\cdot i + \beta_{4}\cdot d + \alpha,
\end{equation}
where $Y$ is a continuous variable measuring the extent of a single frame (e.g., exoticism), the $\beta_{i}$ terms are coefficients, $C$ is categorical cuisine (with US as the reference level), $l$ is review length, $p$ is categorical restaurant price (with \$\$ as the reference level), $s$ is restaurant mean star rating, $i$ is neighborhood income, $d$ is neighborhood diversity, and $\alpha$ is an intercept term capturing the framing score of a review at the reference level for all categorical variables (i.e. a review of a \$\$ US restaurant). %} 
This model predicts each form of othering from cuisine region, while controlling for review length, restaurant mean star rating, restaurant price point, neighborhood median income, and neighborhood racial diversity. Additionally, we qualitatively explore how each cuisine is othered using the Fightin' Words method (see Eq. 1; also \citet{monroe2008fightin}).

We find from our regression analyses that overall, \textbf{H1a-c} are all borne out: othering along all 3 dimensions is quantitatively more prevalent in reviews of immigrant cuisine (Figure \ref{fig:study1_results}). However, we also observe intra-cuisine variance: othering of immigrant cuisine is driven primarily by \textsc{As} and \textsc{Lat}, with both cuisines consistently framed as more exotic, and \textsc{As} framed as more prototypical. In contrast, \textsc{Eur} is not framed as significantly more exotic or prototypical than non-immigrant cuisine (which is consistently associated with a decrease in othering). The negative coefficients for non-immigrant and \textsc{Eur} exoticism in particular support \citeauthor{janer2005cooking}'s \citeyearpar{janer2005cooking} suggestion that US dining culture remains strongly rooted in a Western culinary perspective,\footnote{Or perhaps, a Global North perspective. The Global North vs.~South division could also explain why Japan patterns more like Europe than Asia for certain kinds of framing, though we do not see this for South Korea (see per-cuisine results in the Appendix).} with non-European immigrant cuisines perceived as different with respect to this implicit reference point. %and potentially contributes to its continuing dominance as the default.
%(``This is not to say that the strongly exotic frame is not present in omnivorous food culture, but it does suggest that culinary omnivores are more interested in weakly exotic foods (situated in immigrant American food communities, rural American settings, or Europe) than in strongly exotic foods based in continents outside of Europe and North America. It also supports Janer’s (2005) suggestion that gourmet food culture remains strongly situated in the North American and \textsc{Eur} core, with other global food cultures added intermittently in ways that do not fundamentally challenge a Eurocentric culinary canon'' (Johnston and Baumann (2007: 192))

Qualitatively, we find from applying the Fightin' Words method (Eq. 1) that there are also notable differences in how \textsc{Eur} vs. \textsc{As} and \textsc{Lat} are othered (Table \ref{tab:othering_frames_lor}). Although all cuisines are described with neutral prototypicality features (e.g., \textit{regular}, \textit{standard}, \textit{typical}), only \textsc{Eur} is described with positive features like \textit{classic} and \textit{exemplary}. Both the neutral and positive features represent essentializing language in that they reduce a cuisine to a prototype, but the positive features do so by placing favorable emphasis on certain representative elements of a cuisine. Our results also corroborate \citet{gualtieri2022discriminating}, who found that non-white restaurants in the \textit{Michelin Guide} tend to be described in terms of authenticity, and European restaurants according to what they call the ``logic of technique,'' or terms like \textit{exemplary} that assert the existence of a prized culinary canon.
%Both types of essentializing language are a form of narrowing and erasure of nuance within a cuisine, but the former narrows a cuisine by limiting what it gets to be in the popular imagination, whereas the latter narrows a cuisine by highlighting certain representative dishes.

Further, we notice that authenticity is afforded to \textsc{Eur} through adjectives of simplicity, such as \textit{homemade, rustic,} and \textit{quaint}, whereas \textsc{Lat} and \textsc{As} are afforded authenticity by being \textit{authentic} or \textit{legit}. %\footnote{We fit additional regression models %(controlling for review length, restaurant mean star rating, restaurant price point, neighborhood median income, and neighborhood racial diversity) 
%to predict distinct types of authenticity framing and find that authenticity of \textsc{Eur} is driven mostly by simplicity features, %($\beta$\textsubscript{Eur} (the effect of being a European restaurant)=0.078, p=0.0; $\beta$\textsubscript{Lat}=0.038, p=1.4e-86; $\beta$\textsubscript{As}=-0.053, p=8.0e-86), 
%whereas authenticity of \textsc{As} and \textsc{Lat} is driven by other kinds of authenticity features%($\beta$\textsubscript{Eur}=0.018, p=0.0; $\beta$\textsubscript{Lat}=0.27, p=0.0; $\beta$\textsubscript{As}=-0.099, p=0.0)
%.} 
In particular, \textsc{Eur} authenticity features evoke the current prestige fare of simple, artisanal, farm-to-table cooking championed by restaurateurs such as Alice Waters of Chez Panisse. As \citet[p.100]{ray2007ethnic} writes: ``The craftsmanship of bourgeois home-cooking was the new posture, contrasted with the mannered style of French haute cuisine [...] Rusticity replaced elegance.%That went hand-in-hand with the crusade of Alice Waters\textemdash the owner of Chez Panisse\textemdash on behalf of organic and seasonal produce. She also unlocked the door to the Mediterranean by way of the south of France
'' In other words, the framing of \textsc{Eur} authenticity via simplicity may reflect high cultural capital rather than authenticity per se. 

% \begin{figure}
%     \centering
%     \includegraphics[scale=0.5]{figs/auth_results.pdf}
%     \caption{Linear regression coefficients from modeling the presence of authenticity framing \todo{state in just a sentence} in reviews as a function of cuisine type, showing that authenticity framing of \textsc{Eur} cuisine is driven mostly by simplicity frames, whereas authenticity framing of \textsc{As} and \textsc{Lat} cuisines is driven by other kinds of authenticity framing. Regressions control for review length, restaurant mean star rating, restaurant price point, neighborhood median income, and neighborhood racial diversity. Error bars are on the order of graph point size.}
%     \label{fig:auth_results}
% \end{figure}

%TODO: are \textsc{Eur} cuisines framed with more neophilia and \textsc{As} cuisines with more neophobia? 

\xhdr{Study 1B: Othering by cultural outsiders} Since we posit that the above framing dimensions of exoticism, prototypicality, and authenticity are forms of gastronomic othering, in which a cuisine is objectified by outsiders, we expect othering to interact with the racial background of reviewers. For instance, \citet{boch2021mainstream} found that reviews of Mexican restaurants mentioned authenticity less in areas with larger Mexican populations. Demographic information for Yelp users is not available in the dataset, so we similarly examine the influence of the \% population of self-identifying Asian and Hispanic residents in the neighborhood as a proxy, using 2020 census data linked to each restaurant's postal code.\footnote{\url{https://data.census.gov/}} 
%The distributions of \textsc{As} and Hispanic residents, respectively, for restaurant neighborhoods are shown in Figure \ref{fig:pct_race}. 
In cases where users self-divulge that they are not locals (e.g., ``I'm from out of state''%``I'm visiting from California''
), we detect and exclude those users' reviews with a regex filter. %See Appendix for distributions of \textsc{As} and Hispanic residents, respectively, for restaurant neighborhoods.

\begin{figure}[ht]
    \centering
    \includegraphics[scale=0.42]{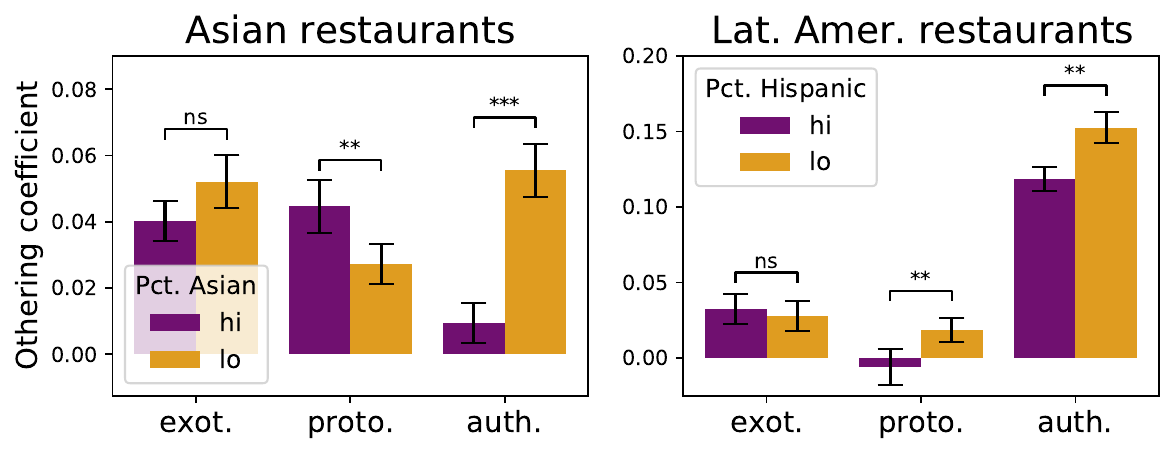}
    \caption{\textbf{Othering by outsiders}: Linear regression coefficients predicting othering of \textsc{As}/ \textsc{Lat} within neighborhoods with a high/low \% (i.e. above/below the median) of Asian %(Mdn.=0.027, IQR=0.026) 
    and Hispanic %(Mdn.=0.13, IQR=0.12) 
    residents.
    %, showing that a lower \% population of the associated race increases authenticity framing of \textsc{As} and \textsc{Lat}, respectively. A lower \% population of the associated race increases prototypicality framing of \textsc{Lat} but \textit{decreases} prototypicality framing of \textsc{As}. Race of residents did not have a significant effect on exoticism framing. 
    Error bars are 95\% CIs. Significance determined by a Wald test comparing coefficients within the same model. ns=p$>$0.05, **=p$<$0.01, ***=p$<$0.001.}
    \label{fig:race_othering_results}
\end{figure}

We fit two separate linear regression models\footnote{
Regressions are fit using the following equation: \begin{equation}
    Y_{j} = \beta_{0}\cdot P_j + \beta_{1}\cdot l + \beta_{2}\cdot p + \beta_{2}\cdot s + \beta_{3}\cdot i + \beta_{4}\cdot d + \alpha,
\end{equation}
where $Y_j$ is a continuous variable measuring the extent of a single frame within either \textsc{As} or \textsc{Lat} (e.g., exoticism of \textsc{As}), the $\beta_{i}$ terms are coefficients, $P_j$ is a categorical coding of the \% self-identifying population of the associated race (\texttt{hi} if \% pop. $\geq$ the median, else \texttt{lo}, with \texttt{hi} as the reference level), $l$ is review length, $p$ is categorical restaurant price (with \$\$ as the reference), $s$ is restaurant mean star rating, $i$ is neighborhood income, $d$ is neighborhood diversity, and $\alpha$ is an intercept term capturing the framing score of a review at the reference level for all categorical variables (i.e. a review of a \$\$ restaurant in a neighborhood with a high \% population of the associated race). Since regressions are fit within the sample of \textsc{As} or \textsc{Lat} restaurants, we omit the cuisine type variable.} to examine the effect of local self-identifying Asian and Hispanic \% population on othering of \textsc{As} and \textsc{Lat}, respectively (i.e. the effect of cultural outsiders). We find that a higher \% population of cultural outsiders significantly increases authenticity framing for \textsc{As} and \textsc{Lat} (Figure \ref{fig:race_othering_results}).
%Hispanic population significantly increases authenticity framing in reviews of \textsc{Lat} restaurants, and similarly, a lower \textsc{As} population significantly increases authenticity framing of \textsc{As} restaurants. 
Interestingly, a higher \% population of outsiders did not have significant effects on exoticism framing and had inconsistent effects on prototypicality framing: while areas with fewer Hispanic residents had more prototypicality framing, those with more Asians had less. %in reviews of \textsc{As} and \textsc{Lat} 
%\footnote{As a sanity check, we fit linear regression models to examine interaction effects between \% of Asian/Hisp. residents (as continuous variables) and cuisine region. We find that a lower \% of Asian/Hisp. residents increases authenticity framing of \textsc{As} and \textsc{Lat} ($\beta$\textsubscript{auth.\textsc{As}} (the effect of \% Asian residents on authenticity framing)=-0.006, p=0.0; $\beta$\textsubscript{auth.Hisp.}=-0.02, p=0.0) and exoticism framing ($\beta$\textsubscript{ex.\textsc{As}}=-0.005, p=0.002; $\beta$\textsubscript{ex.Hisp}=-0.004, p=0.049). A lower \% of Hisp. residents also increases prototypicality framing of \textsc{Lat}, but a lower \% of Asian residents decreases prototypicality framing of \textsc{As} ($\beta$\textsubscript{proto.Hisp.}=-0.02, p=0.0; $\beta$\textsubscript{proto.\textsc{As}}=0.02, p=0.0).} 
However, these results may be skewed by the narrow range of geographic regions represented in our dataset, especially in the case of Asian neighborhoods: 83\% of the neighborhoods with a high Asian \% population are located in Pennsylvania, and 36\% of high Hispanic \% population neighborhoods are located in Florida (despite PA having only 4\% of the overall US Asian population and FL having only 26\% of the overall US Hispanic population). 
%Nonetheless, together with Figure \ref{fig:study1_results}, these results support the finding that authenticity framing is a primary form of othering outsider, while exoticism and prototypicality framing show more variance. 

\subsection{Study 2: Low status framing of non-European cuisines}

To test our second set of hypotheses concerning status framing, i.e. restaurants of less assimilated immigrant groups are reviewed less in high status terms of luxury \textbf{H2a} and more in low status terms of cost \textbf{H2b} and hygiene \textbf{H2c},
we fit linear regression models (Equation \ref{eq:regression}) to predict status framing %(as the dependent variable) 
from cuisine region, while controlling for the same confounds as in Study 1A (restaurant price point, stars, neighborhood income and diversity). Since Study 2 does not make explicit predictions about othering, we did not examine the effect of reviewer racial background as in Study 1B. 
%and consequently controlled only for overall neighborhood diversity. 
To qualitatively explore how different cuisines are framed as high vs.~low status, we used the Fightin' Words method (Eq. 1) to retrieve the features most associated with each cuisine.

\begin{figure}[ht]
\centering
\includegraphics[scale=0.32]{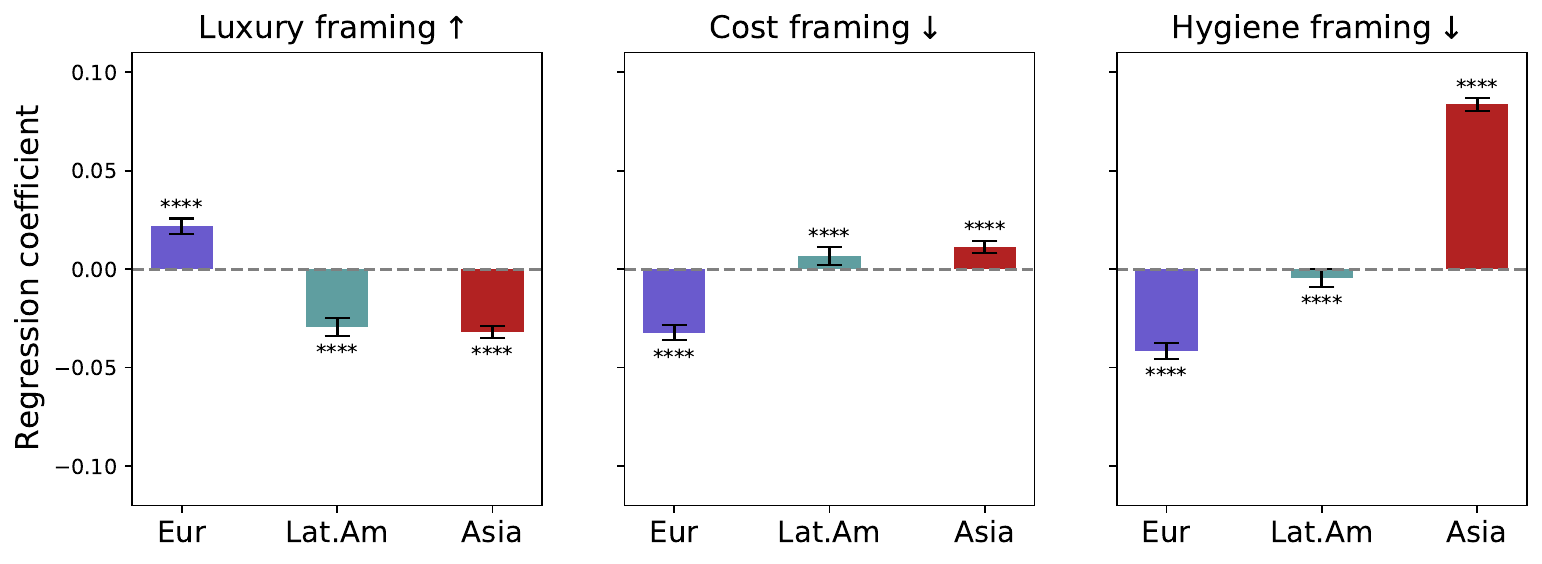}
    \caption{\textbf{Europe afforded higher status}: Linear regression coefficients predicting high $\uparrow$ (luxury) and low status $\downarrow$ (cost; hygiene) framing from cuisine region, controlling for review length, restaurant mean star rating, restaurant price point (reference level: \$\$), neighborhood median income and racial diversity. %\textsc{Eur} receives more high status framing and \textsc{Lat} and \textsc{As} receive more low status framing. 
    Error bars are 95\% CIs. 
    Significance value for \textsc{Eur} (the reference level) indicates whether being \textsc{Eur} has a significant effect on the outcome variable, other values indicate whether a cuisine is significantly different from \textsc{Eur} with respect to its effect on the outcome. ns=p$>$0.05, **=p$<$0.01, ***=p$<$0.001.}
    \label{fig:study2_results}
\end{figure}

We find from our regression analyses support for \textbf{H2}: \textsc{Lat} and \textsc{As} (associated with less assimilated Latin Americans and Asians) receive less %positive sentiment and 
luxury framing, and more cost and hygiene framing, compared to \textsc{Eur} (Figure \ref{fig:study2_results}). 

Qualitatively, we find from the Fightin' Words analysis that %the positive features most associated with \textsc{Eur} are higher intensity (e.g., \textit{wonderful, perfect}) compared to those associated with \textsc{Lat} (e.g., \textit{great, cool}) and with \textsc{As} (e.g., \textit{good, nice}) 
the cost features associated with \textsc{Eur} also tend to connote luxury (\textit{expensive, pricey}) compared to those associated with \textsc{Lat} and \textsc{As} (e.g., \textit{cheap, affordable}; Table~\ref{tab:study2_lor}). %Noticeably, there are no hygiene features significantly associated with \textsc{Eur} relative to \textsc{Lat} and \textsc{As}, suggesting that hygiene is rarely commented on in reviews of \textsc{Eur}. 

\begin{figure}[ht]
    \centering
    \includegraphics[scale=0.4]{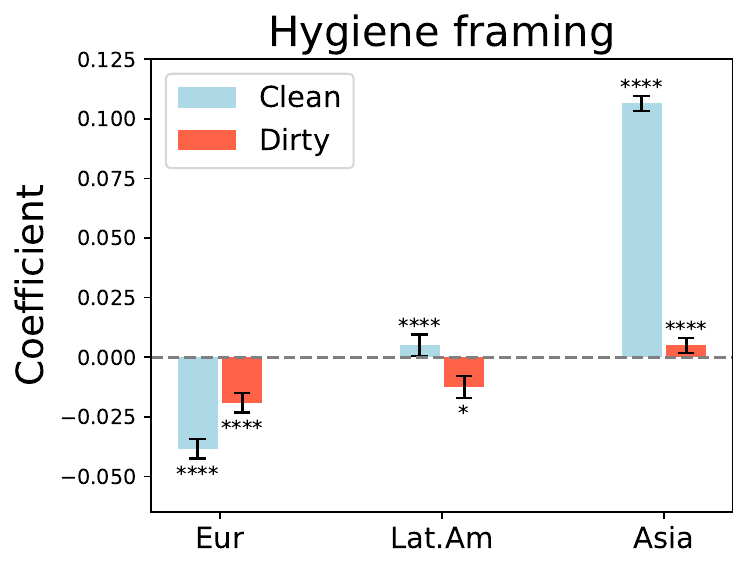}
    \caption{\textbf{\textsc{As} and \textsc{Lat} evaluated more on hygiene}: Linear regression coefficients predicting 
    %$+/-$ hygiene 
    clean, dirty framing from cuisine region, controlling for review length, restaurant mean star rating, restaurant price point (reference level: \$\$), neighborhood median income and racial diversity. %\textsc{As} receives the highest levels of both %$+$ (clean) and $-$ (dirty) 
    %types of hygiene framing.
    Significance value for \textsc{Eur} (the reference level) indicates whether being \textsc{Eur} has a significant effect on the outcome variable, other values indicate whether a cuisine is significantly different from \textsc{Eur} with respect to its effect on the outcome. ns=p$>$0.05, **=p$<$0.01, ***=p$<$0.001.}
    \label{fig:hygiene_results}
\end{figure}

\begin{figure}[ht]
\centering
\includegraphics[scale=0.32]{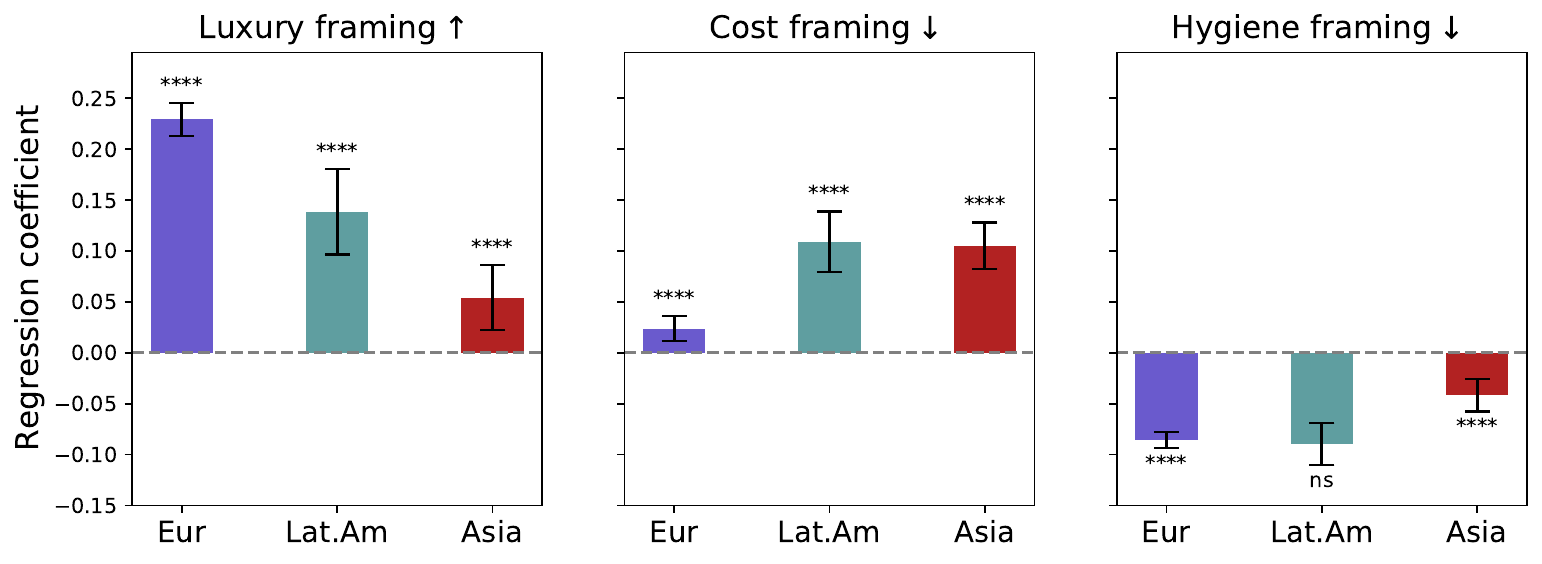}
    \caption{\textbf{Glass ceiling}: Linear regression coefficients predicting high $\uparrow$, low $\downarrow$ status framing %in reviews 
    of \$\$\$-\$\$\$\$ restaurants from cuisine region, controlling for review length, restaurant mean rating, neighborhood income, racial diversity. 
    %\textsc{Eur} receives more high-valuation framing; \textsc{As} receives more low-valuation framing; \textsc{Lat} receives more cost framing (differences in positivity and hygiene framing are not significant). 
    Error bars are 95\% CIs. 
    %Significance value for \textsc{Eur} (the reference level) indicates whether being \textsc{Eur} has a significant effect on the outcome variable, other values indicate whether a cuisine is significantly different from \textsc{Eur} with respect to its effect on the outcome. 
    ns=p$>$0.05, **=p$<$0.01, ***=p$<$0.001.}
    \label{fig:study2b_results}
\end{figure}
%Dirty dining is a pervasive negative stereotype associated with \textsc{As} restaurants~\cite{hirose2011no,lee2012consumer}\textemdash is there evidence for this stereotype on a larger scale, and potentially for \textsc{Lat} restaurants as well? We create feature dictionaries to measure hygiene framing on aggregate, and cleanliness and dirtiness, respectively. We then fit linear regression models to estimate each type of hygiene framing from cuisine region and 
Disaggregating hygiene framing into %positive hygiene (i.e. clean) and negative hygiene (i.e. dirty),
clean and dirty, we find from regression analyses that \textsc{As} continues to receive the most 
%in terms of both types of hygiene 
of both framing types (Figure \ref{fig:hygiene_results}). 
%\textsc{Lat} is framed 
%more in terms of positive hygiene 
%as cleaner than \textsc{Eur}, but not significantly dirtier. %negative hygiene. 
Interestingly, \textsc{Eur} is the only cuisine to receive more dirty than clean framing, suggesting that its cleanliness may be taken for granted, and hygiene conditions are only noteworthy when dirty. 
%Indeed, reviewing a restaurant as \textit{clean} is only necessary if expectations are otherwise. 
Evaluating a restaurant in terms of hygiene, regardless of whether the framing is positive (clean) or negative (dirty),  presupposes that cleanliness is a relevant and important dimension of discussion, reflecting and 
potentially reinforcing negative assumptions about sanitary conditions of \textsc{As} and \textsc{Lat}. %The tendency to evaluate non-\textsc{Eur} cuisines along such terms may be another reflection of reduced cultural capital: instead of appraising a restaurant for its aesthetic or higher order experiential attributes, customers focus on lower order concerns regarding hygiene.

\begin{table*}[ht]
\footnotesize
    \centering
    \begin{tabular}{clll}
\toprule
Frame & Most \textsc{Eur} & Most \textsc{Lat} & Most \textsc{As} \\
\midrule
% Positivity & romantic (19.7) wonderful (17.5) & good (10.7) amazing (4.2) & good (38.3) special (16.1) tender (10.4) \\ 
% & cozy (13.6) excellent (12.8) lovely (11.6) & super (4.0) fun (3.8) & nice (8.5) delicate (7.0) satisfying (5.9) \\ 
% & charming (9.2) perfect (8.5) & great (3.3) awesome (3.0) & calming (5.8) pleasant (5.0) rich (4.5) \\ 
% & fabulous (8.0) rich (7.2) special (6.3) &  & bold (4.1) pleasing (3.5) original (2.9) \\ & divine (5.8) fantastic (5.6) & & pretty (2.8) welcoming (2.5) \\ 
% & gorgeous, flawless, & \\ 
% & heavenly, delectable, & \\ 
% & amazing, great, & \\ 
% & relaxing, scrumptious, & \\ 
% & rich & \\
Luxury & delicate (3.2) elegant (3.0) & outstanding (4.2) & delicate (8.7) sleek (5.9) pleasing (4.6) \\ 
& exquisite (2.8) & & tasteful (4.5) ornate (3.0) posh (2.1) stylish (2.0) \\ 
% Negativity & empty, bitter, & bad, fake \\ 
% & disappointing, rude, & crazy, weird \\ 
% & arrogant, awful, lousy & disgusting, numb \\ 
% & & gross, wicked \\ 
% & & fiery, horrible \\
Cost & expensive (3.7) pricey (2.5) & cheap (5.7) inexpensive (4.0) & cheap (13.2) affordable (8.5) \\ 
& & affordable (2.2) & inexpensive (6.2) \\
Hygiene & \textemdash & clean (5.4) & clean (34.1) stinky (4.9) unhygienic (2.5) \\
\bottomrule
\end{tabular}
    \caption{Lemmas most associated with high, low status frames of each immigrant cuisine. Association strength measured as the z-score of the weighted log odds ratio between a feature occurring in a review of a region over all other regions (see Eq. \ref{eq:log_odds}).}
    \label{tab:study2_lor}
\end{table*}

\xhdr{Glass ceiling effect} Since status is confounded with restaurant price point, we additionally study framing within reviews of \$\$\$ and \$\$\$\$ restaurants (N=166K). We group \$\$\$ with \$\$\$\$ restaurants due to the small number of the former (N=326). Although we control for price point in all analyses, we wished to more carefully examine the extent of framing disparities: even within the most upscale tier, do non-\textsc{Eur} cuisines continue to receive more low status and less high status framing compared to \textsc{Eur}?

We find that framing disparities persist within reviews of high price point restaurants, pointing to a glass ceiling effect by which \textsc{As} and \textsc{Lat} are viewed as lower status than 
their \textsc{Eur} counterparts (Figure \ref{fig:study2b_results}).\footnote{We also examined framing within James Beard award-winning restaurants but found no significant effects.} This effect is most pronounced for \textsc{As}, which continues to be evaluated as %less positive, 
less luxurious, and more in terms of cost and hygiene, and slightly less pronounced for \textsc{Lat}, which continues to be evaluated as less luxurious and more in terms of cost.

\subsection{Study 3: Reporting bias in LLMs}

LLMs are being used increasingly for content creation in all sorts of domains\textemdash marketing, academic research, legal contexts, creative writing\textemdash and by users who may be unaware of the social biases and stereotypes they contain. As a result, the damage from these representational harms can quickly compound as they are reproduced automatically on large scales. Despite debiasing efforts, simple prompts suggest that stereotypes in human reviews may propagate to LLMs. For instance, when asked to compare French and Korean culture, one of the ``key distinctions'' that ChatGPT identifies is ``Cuisine: French cuisine is renowned for its sophistication, emphasis on quality ingredients, and intricate preparation methods.
Korean cuisine is characterized by bold flavors, a variety of side dishes (banchan), and a reliance on staples like rice and kimchi'' (Figure \ref{fig:llm_example}).
\begin{figure}
    \centering
    \fcolorbox{black}{Gray}{\begin{minipage}{.45\textwidth}
    %\textbf{(Chinese)} 
    \textcolor{white}{\textsf{French and Korean cultures exhibit significant differences shaped by their unique histories, societal norms, and traditions. Here are some key distinctions: 1. Cultural Orientation: French culture tends to be more individualistic [...] 5. Cuisine: French cuisine is renowned for its sophistication, emphasis on quality ingredients, and intricate preparation methods. Korean cuisine is characterized by bold flavors, a variety of side dishes (banchan), and a reliance on staples like rice and kimchi [...]}}
    \end{minipage}}
    \caption{ChatGPT's response to the prompt: ``What are some differences between French and Korean culture?'' Retrieved December 26, 2023.}
    \label{fig:llm_example}
\end{figure}

%being used by children, in essay-writing, etc. etc. all kinds of domains. List a bunch of downstream harms. It would be terrible if biases are propagating. Simple prompts suggests biases propagate. Let’s verify this. To quantitatively check how they propagate, we need experiment. 

%Framing disparities in Yelp reviews matter not only because they may affect consumer perceptions and decision-making, but also because they may propagate to LLMs trained on online text. For instance, when asked to compare French and Mexican restaurants, 

Here we test our final hypothesis \textbf{H3}, whether framing disparities in online text corpora indeed propagate to the LLMs trained on them. To do this, we conduct a controlled review-generation task in which we prompt GPT-3.5 to write reviews evaluating hypothetical restaurants that differ in price point and cuisine using a variety of prompts and models (Appendix Table \ref{tab:llm_prompts}). We then quantitatively test \textbf{H3} using regression models\footnote{Regressions are fit using the following equation: \begin{equation}\label{eq:llm_regression}
    Y = \beta_{0}\cdot C + \beta_{1}\cdot s + \alpha,
\end{equation}
where $Y$ is a continuous variable measuring the extent of a single frame (e.g., exoticism), the $\beta_{i}$ terms are coefficients, $C$ is categorical cuisine (with US as the reference level), $s$ is review sentiment (with \texttt{neutral} as the reference), and $\alpha$ is an intercept term capturing the framing score of a review at the reference level for all categorical variables (i.e. a neutral review of a US restaurant). We do not control for review length or restaurant price point as we balanced synthetic reviews across these parameters for all cuisine and sentiment combinations.} to predict othering and status framing from cuisine region. We also qualitatively study LLM stereotypes using the Fightin' Words method (Eq. 1). 

Regardless of the specific prompt and model used,\footnote{Results presented in this section are based on reviews generated by \texttt{gpt-3.5-turbo-0613} using the prompt, \textit{A customer posted the following restaurant review to an online restaurant review website:$<$span class=`headline' title=``\textbf{[sentiment]} review about a \textbf{[price point]} restaurant, focused on the \textbf{[focus]},''} with the fields in bold varied to mimic Yelp reviews. We obtained similar results using other prompts and model combinations (see Appendix).} we find from our regression analyses (N=58K) that GPT-3.5 exhibits many of the same tendencies as Yelp consumers, such as othering \textsc{As} and \textsc{Lat} through frames of authenticity and exoticism (Table \ref{tab:LLM_othering}). However, we also see that immigrant restaurants are framed as less prototypical compared to non-immigrant restaurants. Nevertheless, closer inspection reveals that the main overlap between synthetic reviews and our prototypicality lexicon is due to the word \textit{classic} (rather than words like \textit{stereotypical} or \textit{standard}, as we found in human reviews), suggesting that our measurement of prototypicality captures a construct more akin to ``iconicity'' in the context of synthetic reviews. Moreover, similar to human reviews, we found that LLMs frame \textsc{Eur} with higher status 
compared to \textsc{As} and \textsc{Lat} (Table \ref{tab:LLM_prestige}). 

Finally, qualitative results of the features most strongly associated with each region (Table \ref{tab:study3_lor}) exemplify additional stereotypes, such as the stereotype of tropicalism for \textsc{Lat} (e.g., \textit{vibrant, lively, festive})~\cite{martynuska2016exotic}, suggesting that gastronomic and cultural stereotypes in LLMs are not limited to the ones we study in the present work.

\begin{table}[ht]
\footnotesize
    \centering
    \begin{tabular}{clll|l}
\toprule
% {} & & & & Immigrant \\
{} &  \textsc{Eur} & \textsc{Lat} &    \textsc{As} &   Immigrant \\
\midrule
Exot. &  -0.01 &         { }\textbf{0.09**} &   { }\textbf{0.07*} &   { }0.05 \\
Proto.  &  \textbf{-0.09**} &        \textbf{-0.14***} &  \textbf{-0.23***} &  \textbf{-0.15***} \\
Auth.   &   { }\textbf{0.32***} &         { }\textbf{0.33***} &   { }\textbf{0.30***} &   { }\textbf{0.32***} \\
\bottomrule
\end{tabular}
    \caption{\textbf{Othering by LLMs}: Linear regression coefficients predicting othering in GPT-3.5 reviews from cuisine region, controlling for review sentiment (reference level: neutral), showing that, as with Yelp reviews, immigrant restaurants are othered with frames of authenticity, and \textsc{As} and \textsc{Lat} with frames of exoticism. Unlike actual reviews, immigrant restaurants are framed as less prototypical. Each row represents a separate regression. *=p$<$0.05, **=p$<$0.01, ***=p$<$0.001.}
    \label{tab:LLM_othering}
\end{table}

\begin{table}[ht]
\footnotesize
    \centering
    \begin{tabular}{clll}
\toprule
{} &       \textsc{Eur} & \textsc{Lat} &         \textsc{As} \\
\midrule
% Positivity  &     { }\textbf{0.13*} &   \textbf{-0.13***} &  \textbf{-0.08***} \\
Luxury         &  { }0.02 &   \textbf{-0.13***} &    { }0.06 \\
Hygiene (aggregate)         &  \textbf{-0.18***} &       -0.14 &  { }\textbf{0.04***} \\
Hygiene (clean)     &  \textbf{-0.15***} &       -0.09 &   { }\textbf{0.09***} \\
Hygiene (dirty)     &    \textbf{-0.10*} &       -0.11 &     \textbf{-0.03*} \\
Cost (aggregate)       &       \textbf{-0.10*} &        -0.11 &       -0.11 \\
Cost (expensive)   &    -0.06 &        -0.09 &      { }\textbf{0.15**} \\
Cost (cheap) &      -0.08 &       -0.06 &      { }\textbf{0.02**} \\
\bottomrule
\end{tabular}
    \caption{\textbf{Status in LLMs}: Linear regression coefficients predicting high and low status framing in GPT-3.5-generated reviews from cuisine region, showing %non-European restaurants are framed as less positive, 
    \textsc{Lat} is framed as less luxurious and \textsc{As} more in terms of cost and hygiene. Each row represents a separate regression predicting framing score from cuisine region, with \textsc{Eur} as the intercept and an additional factor of sentiment (reference level: neutral). %, resulting in mostly negative coeff.s since sentiment skews positive). 
    *= p$<$0.05, ** = p$<$0.01, *** = p$<$0.001.}
    \label{tab:LLM_prestige}
\end{table}

\begin{table*}[ht]
\footnotesize
    \centering
    \begin{tabular}{c l} 
    \toprule 
    Region & Dominant features \\ \midrule
         US & southern (18.5) fried (9.8) soulful (5.1) rustic (4.9) crispy (4.4) cheesy (4.4) slow (2.9) classic (2.5) \\ & gooey (2.5) dry (2.5) mediocre (2.4) terrible (2.2) local (2.2) creamy (2.1) overpriced (2.1) \\ 
         \textsc{Eur} & romantic (5.8) charming (4.9) traditional (4.8) hearty (4.1) mashed (3.5) creamy (3.2) perfect (3.1) cozy (2.9) \\ & warm (2.6) red (2.4) stuffy (2.1) soft (2.1) exquisite (2.0) stuffed (2.0) rich (2.0) thin (2.0) \\ 
         \textsc{Lat} & vibrant (11.2) lively (11.2) colorful (6.7) authentic (4.7) black (3.6) energetic (3.4) fun (3.4) shredded (3.2) festive (3.1) \\ & moist (2.6) upbeat (2.4) homemade (2.3) seasoned (2.2) mixed (2.1) tender (2.1) generous (2.0) friendly (2.0) \\ 
         \textsc{As} & wide (6.0) fresh (5.0) aromatic (4.9) clean (4.1) modern (4.1) sticky (3.9) iced (3.7) fragrant (3.5) spicy (3.5) \\ & beautiful (3.2) authentic (3.2) elegant (3.2) steamed (3.2) serene (3.1) hot (2.9) soothing (2.6) helpful (2.6) peaceful (2.5) \\ & balanced (2.4) comfortable (2.3) marinated (2.3) pickled (2.2) light (2.1) quick (2.1) traditional (2.1) stunning (2.1) \\ \bottomrule
    \end{tabular}
    \caption{Features most associated with each cuisine within GPT-3.5-generated reviews. Association strength measured as the z-score of the weighted log odds ratio between a feature occurring in a review of a given region over all other regions.}
    \label{tab:study3_lor}
\end{table*}

% \begin{figure}
%     \centering
%     \includegraphics[scale=0.33]{figs/LLM_study1.pdf}
%     \caption{othering framing in GPT3-generated reviews, showing}
%     \label{fig:gpt_study1_res}
% \end{figure}

% \begin{figure}
%     \centering
%     \includegraphics[scale=0.35]{figs/LLM_study2.pdf}
%     \caption{Sentiment and cultural capital disparities in GPT3-generated reviews}
%     \label{fig:gpt_study2_res}
% \end{figure}

% \begin{figure}
%     \centering
%     \includegraphics[scale=0.35]{figs/LLM_hygiene_framing.pdf}
%     \caption{}
%     \label{fig:gpt_hygiene}
% \end{figure}

% \begin{figure}
%     \centering
%     \includegraphics[scale=0.35]{figs/LLM_cost_framing.pdf}
%     \caption{}
%     \label{fig:gpt_cost}
% \end{figure}

\section{Discussion}

In this work, we examined the framing of immigrant and non-immigrant cuisines within 2.1M Yelp reviews of restaurants located in 14 US states. Despite the reputation of the US as a culinary melting pot, we find that immigrant restaurants are %systematically othered through the socially constructed frame of authenticity, 
systematically described as more authentic, 
and Asian and Latin American restaurants in particular  are described as more exotic. Further, we find that restaurants associated with less assimilated Asian and Latin American immigrants are more likely to be reviewed as cheap and dirty compared to European restaurants. Our results are robust to differences in restaurant price, average star rating, and neighborhood income and racial diversity. We further show that LLMs replicate similar framing tendencies. 

%those associated with less assimilated immigrant groups are systematically framed as %less positive and 
%lower status.

%evidence for systematic othering of immigrant cuisines through socially constructed frames of exoticism, prototypicality, and authenticity. 

%We found a number of framing disparities  non-American restaurants, especially \textsc{As} and \textsc{Lat} ones, are subject to othering in the form of exoticizing and stereotyping language that reinforces a Western culinary default and a prototypical notion of these cuisines. We also corroborated the theory of ethnic succession of taste, showing how \textsc{Eur} cuisine is more associated with high status frames of positive sentiment and luxury, while \textsc{Lat} and \textsc{As} cuisines are more associated with low status frames of affordability and hygiene. 

Beyond creating representational harms, such as reinforcing negative perceptions of immigrant cultures, framing differences in online restaurant reviews may contribute to economic inequalities, since reviews stand to influence the decisions of millions of consumers. Indeed, prior work has shown associations between restaurant review scores and revenue \citep{luca2016reviews}. Future work could seek to better understand and quantify the link between differences in review language and restaurant revenue. 

There are a number of limitations in our current studies that future work could address. We found an inconsistent effect of reviewer race on othering, potentially due to the coarse estimate of race (based on restaurant zipcode) we used. Future research can explore datasets in which racial background at the level of individual users is available.
%Future work can also control for restaurant quality beyond average star rating and price point. 
We also made a number of simplifying assumptions in locating individual cuisines within broader %culinary 
regions, and neglected entire regions due to %data 
sparsity. Future work could explore within-region and within-cuisine variance more carefully, as well as study a wider range of cuisines. In addition, future work could study a more representative sample of US cities, as well as review language outside the US. 

Finally, our work points to a need for strategies to combat representational harms in online food discourse. Controlled psychological experiments could reveal the deeper mechanisms underlying observed framing differences, and AI-mediated tools could intervene during the review-writing process. For instance, web plug-ins could alert reviewers to harmful linguistic choices that they may not have been aware of and provide suggestions for rephrasing. 

\section{Broader perspective}

All review data provided by the Yelp Academic Dataset is posted publicly and anonymized. We perform all our analyses on aggregate and avoid targeting any specific users or restaurants. We rely on restaurants' self-declared cuisine tags rather than inferring cuisine labels from other attributes, which may encourage biased profiling practices. We also use purely geographic criteria to locate individual cuisines within broader regions, so as to limit similar profiling biases.

Due to data sparsity, we regrettably omitted cuisines from Africa, the Middle East, and Oceania, in addition to many individual countries and culinary traditions. We also excluded intersectional cuisines with connections to multiple regions, such as Caribbean cuisine and Tex-Mex. In next steps, we aim to re-include omitted cuisines to resist tendencies of marginalizing minorities. We also recognize that the Yelp user base skews toward college-educated people with high incomes.\footnote{https://www.yelp-press.com/company/fast-facts/default.aspx} Our findings should therefore not be taken as representative of US consumers as a whole. 

Additionally, our experiments with LLMs may engender risks associated with adversarial use (e.g., abuse of our prompting procedure to ``review bomb'' 
%\footnote{https://www.seattletimes.com/business/how-review-bombing-can-tank-a-book-before-its-published/} 
businesses with negative reviews). Finally, by studying stereotypes and frames of cultural prejudice, we risk further reifying these constructs. At the same time, we hope this work stimulates reflection, conversation, and additional research to interrogate and dismantle the preconceived attitudes that each of us as a consumer and restaurant-goer may hold.

\section{Further related work}

Our work extends theoretical and empirical work on the sociology of taste~\cite{bourdieu1987distinction,peterson1997rise,johnston2007democracy,williamson2009working}, especially as it relates to racialized criticism and
the devaluation of marginalized producers~\cite{grazian2005blue,childress2019encultured,chong2011reading,janer2005cooking,liu2009chop,ray2007ethnic,ray2017bringing,gualtieri2022discriminating,hammelman2023paving}. We also draw from the broader literatures on othering, authenticity, and stereotyping~\cite{said1978introduction,martynuska2016exotic} and work examining these attitudes in Yelp reviews~\cite{hirose2011no,kovacs2014authenticity,gottlieb2015dirty,boch2021mainstream} and food contexts more broadly~\cite{freedman2011authenticity,lee2012consumer,oleschuk2017foodies}. We are also inspired by work on representational harms in discourses of race and ethnicity~\cite{card2022computational} and in machines~\cite{crawford2017bias,barocas2017fairness,blodgett2020language,cheng2023marked}. 

Analyzing broader social aspects of food through online text has been an active area of research. Previous work leveraged various data sources, e.g., menus~\cite{jurafsky2016linguistic,turnwald2020five}, search engine logs \cite{West:2013:CCI:2488388.2488510,gligoric2022population}, recipes \cite{wagner2014nature,trattner2018predictability}, and review platforms \cite{jurafsky2014narrative,chorley2016pub}. %, %crowdsourcing platforms \cite{Howell:2016:ATP:2896338.2896358,dunford2014foodswitch}, 
%and geo-located signals \cite{sadilek2018machine}. 
%Given the popularity and prevalence of food-related content, 
ICWSM research, in particular, has focused on studying diets through social media, such as Instagram \cite{garimella2016social,ofli2017saki} and Twitter \cite{mejova2016fetishizing,gligoric2022biased}.
Our work is also related to studies of the impact of reviews on restaurant performance~\cite{luca2016reviews,kim2016impact,wang2021financial} and suggests directions for finer-grained impacts that may affect ethnic groups differentially.

\section{Acknowledgments}

We thank the reviewers and members of the Jurafsky Lab for helpful feedback. Kristina Gligorić is supported by the Swiss National Science Foundation (Grant P500PT-211127). This work is also funded by the Hoffman–Yee Research Grants Program and the Stanford Institute for Human-Centered Artificial Intelligence.

\bibliography{aaai22}

\section{Appendix}\label{sec:appendix}

% \begin{figure}[ht]
%     \centering
%     \includegraphics[scale=0.45]{figs/top_cuisines_filtered_heatmap.pdf}
%     \caption{Heatmap showing normalized overlap among the 20 cuisine labels in our dataset. Axis order reflects decreasing frequency rank. Only restaurants with cuisines associated with a single region are included in our analyses.}
%     \label{fig:cuisine_heatmap}
% \end{figure}

% \begin{figure}[ht]
%     \centering
%     \includegraphics[scale=0.45]{figs/pct_race.pdf}
%     \caption{Distribution of restaurants over percentage of \textsc{As}- (\textbf{A}) and Hispanic-identifying residents (\textbf{B}).}
%     \label{fig:pct_race}
% \end{figure}

\xhdr{LLM prompting details}\label{ssec:prompt_details} We share our prompt templates and model parameters for reproducibility: 
\begin{table}[ht]
    \centering
    \begin{tabular}{l} \toprule
         \textit{1. A customer posted the following restaurant review to}\\ \textit{an online restaurant review website:$<$span class=`head-}\\ \textit{`line' title=``[\textbf{sentiment}] review about a [\textbf{price point}]}\\\textit{[\textbf{cuisine}] restaurant, focused on the [\textbf{focus}]"$>$}\\
         2. \textit{Write a [\textbf{sentiment}] review of a [\textbf{price point}]}\\\textit{[\textbf{cuisine}] restaurant, focusing on the [\textbf{focus}]}
         \\
         3. \textit{Give an example of a [\textbf{sentiment}] review of a}\\\textit{[\textbf{price point}] [\textbf{cuisine}] restaurant} \\ \bottomrule
    \end{tabular}
    \caption{LLM prompt templates}
    \label{tab:llm_prompts}
\end{table}
% \noindent \textit{{A customer posted the following restaurant review to an online restaurant review website: $<$span class=``headline" title=``[\textbf{sentiment}] review about a [\textbf{price point}] [\textbf{cuisine}] restaurant, focused on the [\textbf{aspect in focus}]"$>$}} %\noindent 
For the first 2 of the templates in Table \ref{tab:llm_prompts}, we varied the \textbf{focus}: staff, waitstaff, employees, waiter, waitress, food, drinks, main courses, appetizers, desserts, place, spot, atmosphere, experience, ambiance; for all templates we varied: \textbf{sentiment}: very positive, positive, neutral, negative, very negative; \textbf{price point}: \$ (\$10 and under),
    \$\$ (\$10-\$25),
    \$\$\$ (\$25-\$45),
    \$\$\$\$ (\$50 and up); 
    \textbf{cuisine}: see Table \ref{tab:cuisine_stats}. We use the following parameters: temperature=1, max\_tokens=256, top\_p=1, frequency\_penalty=0, presence\_penalty=0. 

\xhdr{Abridged LLM replication results} 
For robustness, we use various standard prompt types (structured role-playing; instructing to write; instructing to give an example) to prompt \texttt{gpt-3.5-turbo-0613} and the newer \texttt{gpt-3.5-turbo-1106}, which has an expanded context window of 16K tokens (compared to 4K). 
We find that results with different prompts and models  replicate our main results from Study 3 (see Tables \ref{tab:LLM_othering_rep}-\ref{tab:LLM_prestige_rep}; we display a subset of model/prompt combinations due to space limitations). 

\begin{table}[ht]
\footnotesize
    \centering
    \begin{tabular}{clll|l}
\toprule
% {} & & & & Immigrant \\
{} &  \textsc{Eur} & \textsc{Lat} &    \textsc{As} &   Immigrant \\
\midrule
Exot. &  -0.01 &         { }\textbf{0.12**} &   { }\textbf{0.08*} &   { }0.06 \\
Proto.  &  { }0.02 &        \textbf{{-}0.09*} &  \textbf{{-}0.21***} &  \textbf{{-}0.09**} \\
Auth.   &   \textbf{{ }0.29***} &         { }\textbf{0.29***} &   { }\textbf{0.28***} &   { }\textbf{0.29***} \\
\bottomrule
\end{tabular}
    \caption{\textbf{Othering by LLMs}: Linear regression coefficients predicting othering in reviews generated by \texttt{gpt-3.5-turbo-1106} using prompt 1, from cuisine region, controlling for review sentiment (reference level: neutral), showing that immigrant restaurants are othered with frames of authenticity, and \textsc{As} and \textsc{Lat} with frames of exoticism. Unlike actual reviews, immigrant restaurants are framed as less prototypical. Each row represents a separate regression. *=p$<$0.05, **=p$<$0.01, ***=p$<$0.001.}
    \label{tab:LLM_othering_rep}
\end{table}

\begin{table}[ht]
\footnotesize
    \centering
    \begin{tabular}{clll}
\toprule
{} &       \textsc{Eur} & \textsc{Lat} &         \textsc{As} \\
\midrule
% Positivity  &     \textbf{{ }0.29***} &   \textbf{-0.22***} &  \textbf{-0.11**} \\
Luxury         &  { }0.00 &   \textbf{-0.21***} &   { }0.00 \\
Hygiene (aggregate)     &  \textbf{-0.14**} &       { }0.01 &   \textbf{{ }0.11**} \\
Hygiene (clean)     &  {-}0.08 &       { }0.00 &   \textbf{{ }0.13**} \\
Hygiene (dirty)     &    -0.12 &       { }0.02 &     { }0.02 \\
Cost (aggregate)       &       -0.04 &        { }0.04 &      { }0.01 \\
Cost (expensive)   &    \textbf{-0.15**} &        { }0.00 &     {-}0.05 \\
Cost (cheap) &      { }0.07 &       { }0.05 &      { }0.06 \\
\bottomrule
\end{tabular}
    \caption{\textbf{Status in LLMs}: Linear regression coefficients predicting high and low status framing in reviews generated by \texttt{gpt-3.5-turbo-0613} using prompt 2, from cuisine region, showing %non-European restaurants are framed as less positive, 
    \textsc{Lat} is framed as less luxurious and \textsc{As} more in low status terms of hygiene. Each row represents a separate regression predicting framing score from cuisine region, with \textsc{Eur} as the intercept and an additional factor of sentiment (reference level: neutral). *=p$<$0.05, **=p$<$0.01, ***=p$<$0.001.}
    \label{tab:LLM_prestige_rep}
\end{table}

\xhdr{Lexicons}\label{sec:full_lexicons}
\textit{\textbf{Exoticism}} abnormal, bizarre, different, distinct, distinctive, exotic, fascinating, foreign, intriguing, odd, peculiar, strange, unfamiliar, unnatural, unsettling, unusual, weird \textit{\textbf{Protoypicality}} archetypal, archetype, average, basic, characteristic, classic, classical, common, commonplace, definitive, emblematic, essential, everyday, exemplary, generic, habitual, mundane, norm, normal, ordinary, predictable, quintessential, regular, standard, stereotypical, typical, unremarkable, usual \textit{\textbf{Authenticity}} accurate, authentic, hand-made, handmade, home-made, homemade, homey, humble, idyllic, laid-back, laidback, legit, legitimate, modest, original, pastoral, proper, quaint, real, real deal, rural, rustic, simple, traditional, true, unassuming, uncomplicated, unfussy, unpretentious 
% \textit{\textbf{Authenticity-simplicity}} hand-made, handmade, home-made, homemade, homey, humble, idyllic, laid-back, laidback, modest, pastoral, quaint, rural, rustic, simple, unassuming, uncomplicated, unfussy, unpretentious
% \textit{\textbf{Positivity}} keen, terrific, beloved, gorgeous, delighful, attractive, superior, pretty, likeable, fabulous, romantic, tender, wonderful, original, awesome, beautiful, bold, glamourous, nice, fantastic, grand, charming, fab, divine, delicate, ideal, perfect, pleasing, aok, precious, cute, inspiring, virtuosic, fun, jolly, delicious, agreeable, pleasant, graceful, splendid, brilliant, good, impressive, great, merry, magnificent, super, elegant, delectable, rich, optimal, scrumptious, fine, friendly, excellent, satisfying, flawless, wow, amazing, pleasing, sentimental, positive, cool, heavenly, creative, exciting, playful, special, lovely, dynamic, pleasurable, cosy, cozy, inviting, welcoming, calming, relaxing 
\textit{\textbf{Luxury}} alluring, astonishing, breathtaking, classy, dazzling, delicate, dignified, elaborate, elegant, enchanting, enticing, exquisite, extraordinary, extravagant, fashionable, glamorous, glorious, graceful, grand, lavish, lush, luxurious, magnificent, majestic, marvelous, ornate, outstanding, picturesque, pleasing, polished, posh, refined, regal, remarkable, sleek, sophisticated, spectacular, stylish, tasteful, voluptuous \textit{\textbf{Cost}} affordable, bargain, budget, cheap, costly, economical, exorbitant, expensive, inexpensive, low-cost, low-priced, low cost, low priced, overpriced, pricey, uncostly, unexpensive \textit{\textbf{Hygiene}} clean, dirty, disgusting, filthy, grimy, gross, (un)hygienic, messy, nasty, (un)sanitary, smelly, spotless, stinking, stinky, tidy

\begin{figure}[h!]
    \centering
    \includegraphics[scale=0.3]{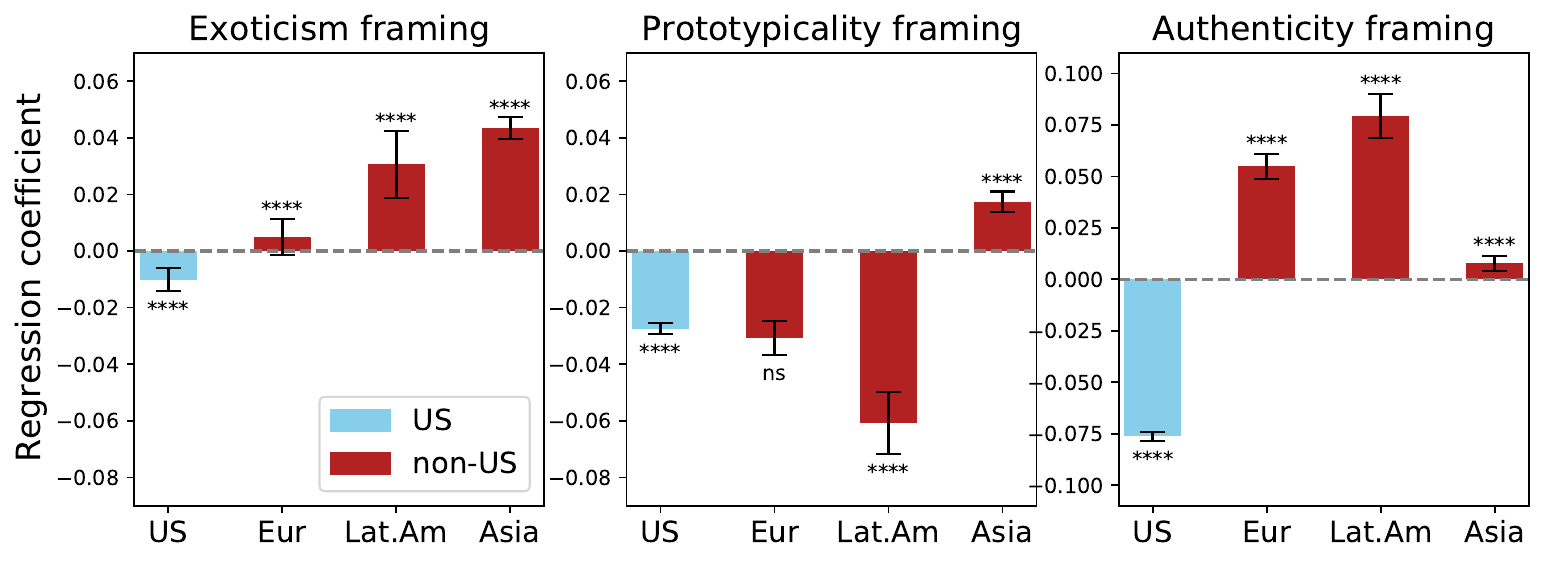}
    \caption{Coeff. from linear regressions estimating othering from cuisine, with \textbf{most frequent cuisine per region removed} (\textit{US: Trad.Am.; Eur: Ital.; Lat: Mex.; As: Chinese}).}
    \label{fig:top_removed_study1_results}
\end{figure}

\begin{figure}[h!]
    \centering
    \includegraphics[scale=0.32]{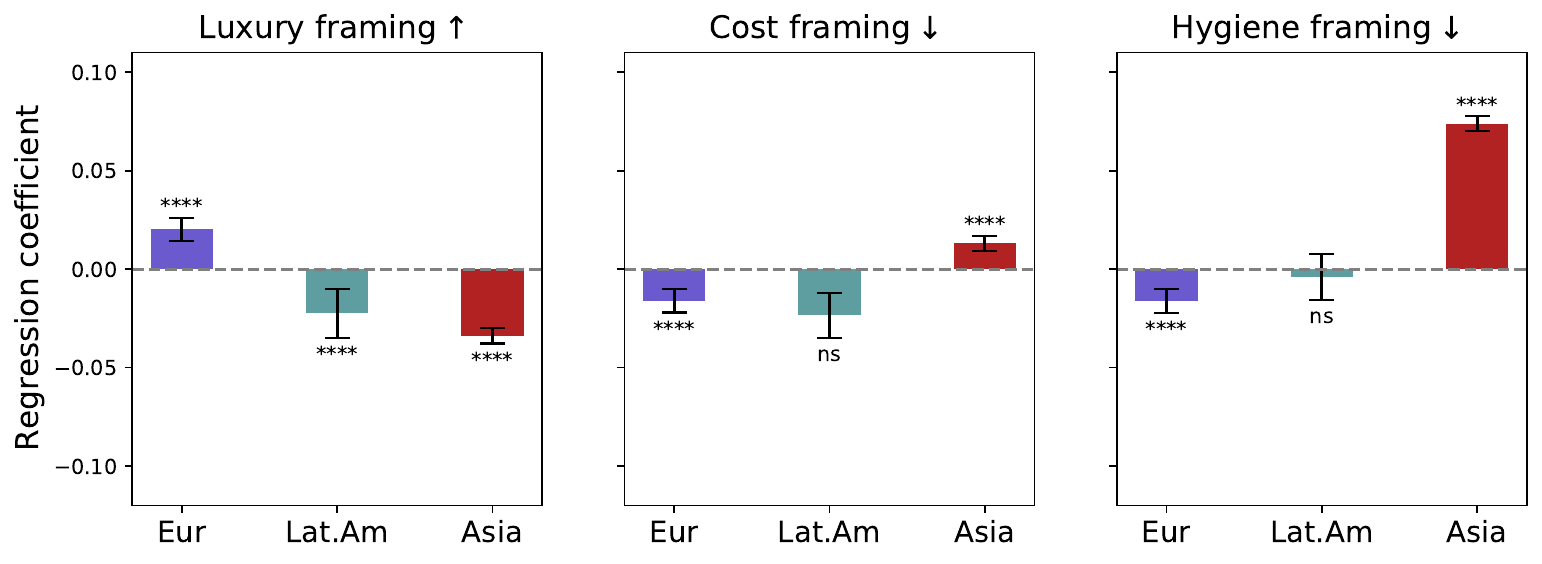}
    \caption{Coefficients from linear regressions estimating high $\uparrow$ vs. low status $\downarrow$ framing from cuisine, with \textbf{most frequent cuisines per region removed}.}% (\textit{US: Trad.Am.; Eur.: Ital.; Lat.Am.: Mex.; Asia: Chinese}).}
\label{fig:top_removed_study2_results}
\end{figure}

\begin{figure}[h!]
    \centering
\includegraphics[scale=0.18]{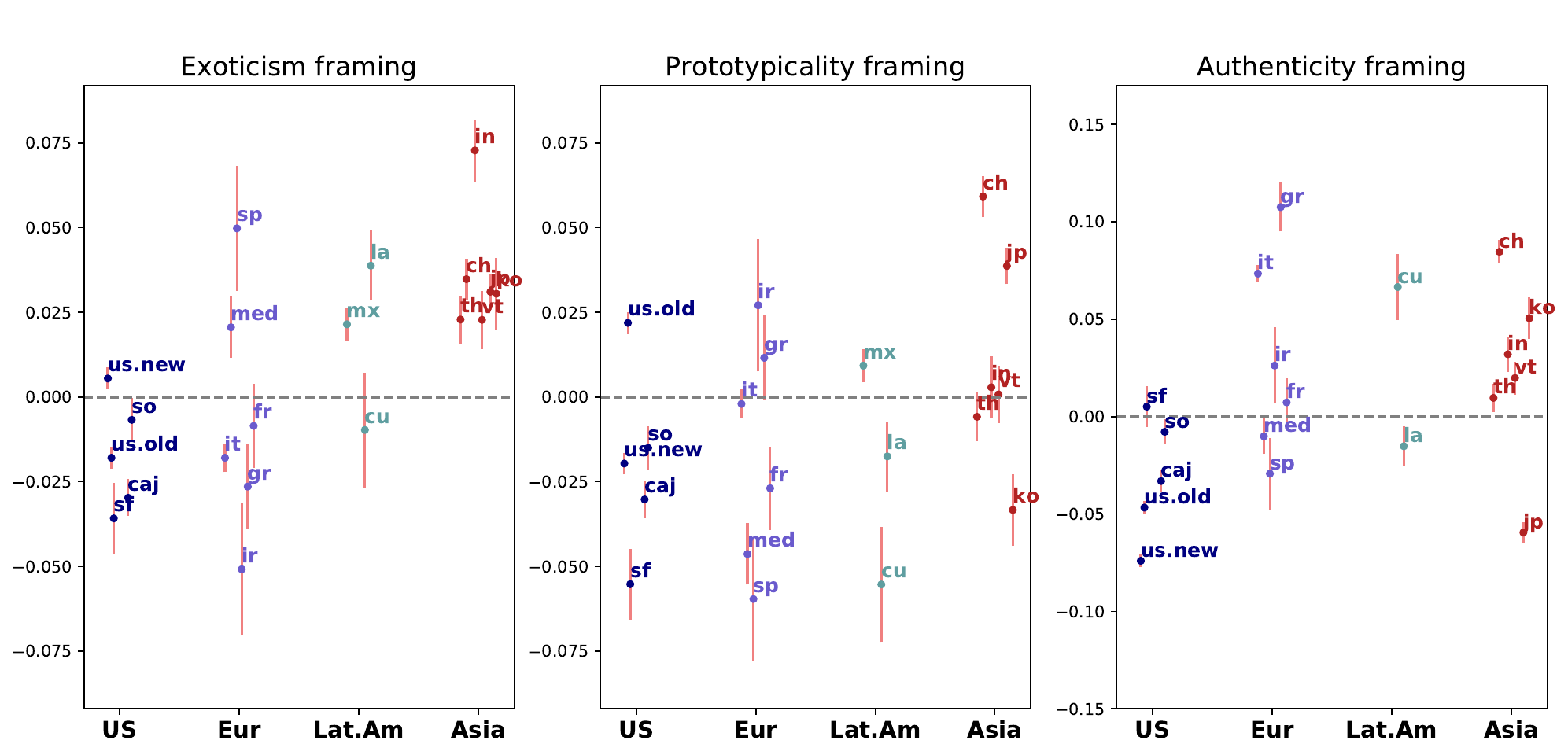}
    \caption{\textbf{Per cuisine coefficients} predicting othering.}
    \label{fig:study1_per_cuisine}
\end{figure}

\begin{figure}[h!]
    \centering
\includegraphics[scale=0.2]{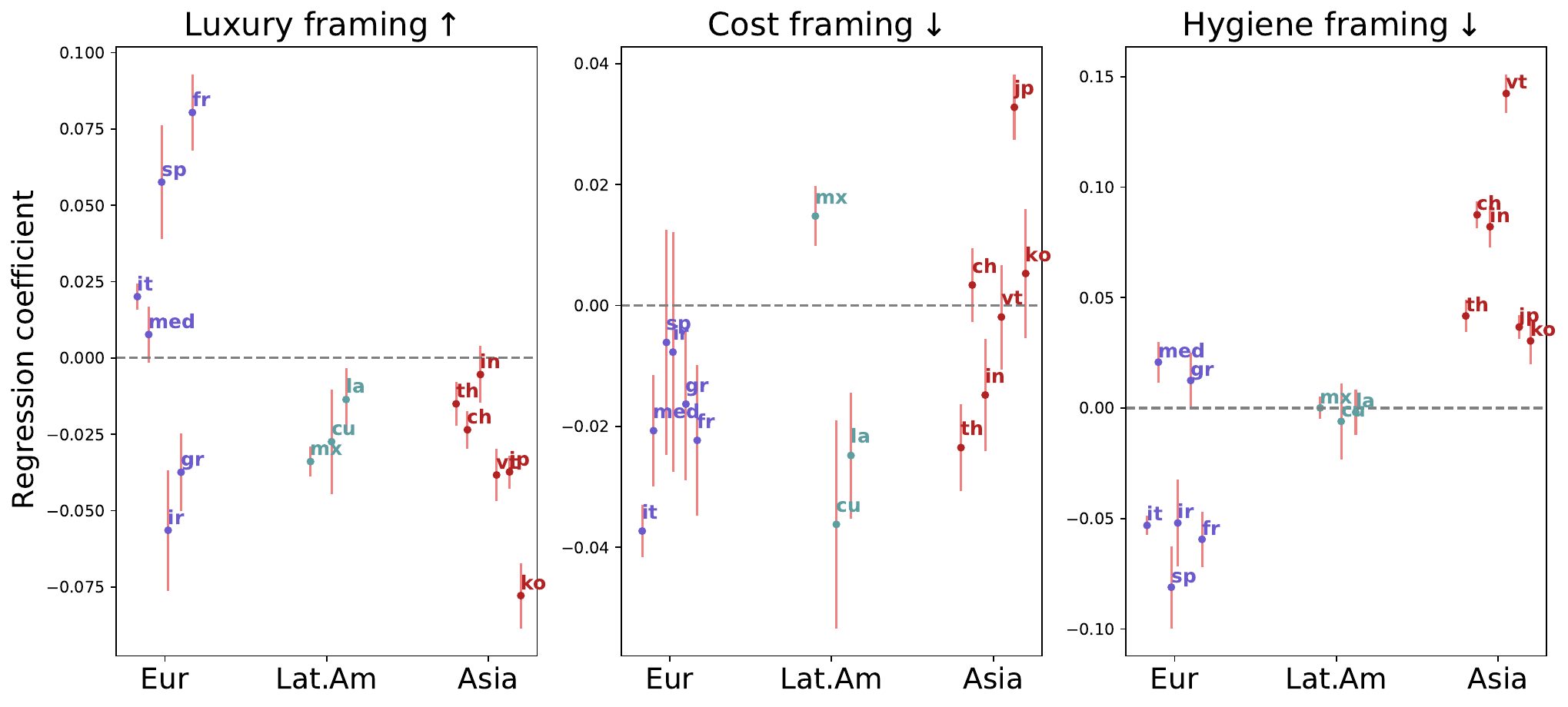}
    \caption{\textbf{Per cuisine coefficients} predicting high $\uparrow$ \& low status $\downarrow$ framing.}
    \label{fig:study2_per_cuisine}
\end{figure}

% \subsection{Per-cuisine analyses}\label{ssec:per_cuisine_results}

% \begin{figure*}[ht]
%     \centering
% \includegraphics[scale=0.4]{figs/glass_ceiling_per_cuisine_results.pdf}
%     \caption{\textbf{Per cuisine regression coefficients} from predicting high $\uparrow$ (positivity; luxury) vs. low status $\downarrow$ (cost; hygiene) framing within \$\$\$-\$\$\$\$ restaurants.}
%     \label{fig:glass_ceiling_per_cuisine}
% \end{figure*}

% \begin{figure}[ht]
%     \centering
% \includegraphics[scale=0.35]{figs/study2_per_cuisine_hygiene_results.pdf}
%     \caption{\textbf{Per cuisine regression coefficients} from predicting positive (clean) and negative (dirty) hygiene framing.}
%     \label{fig:study2_per_cuisine_hygiene}
% \end{figure}

% \begin{figure*}[ht]
%     \centering
% \includegraphics[scale=0.45]{figs/study2_per_cuisine_cost_results.pdf}
%     \caption{Per cuisine regression coefficients from predicting aggregate cost framing and expensive and cheap framing.}
%     \label{fig:study2_per_cuisine_cost}
% \end{figure*}

\begin{figure}[h!]
    \centering
    \includegraphics[scale=0.27]{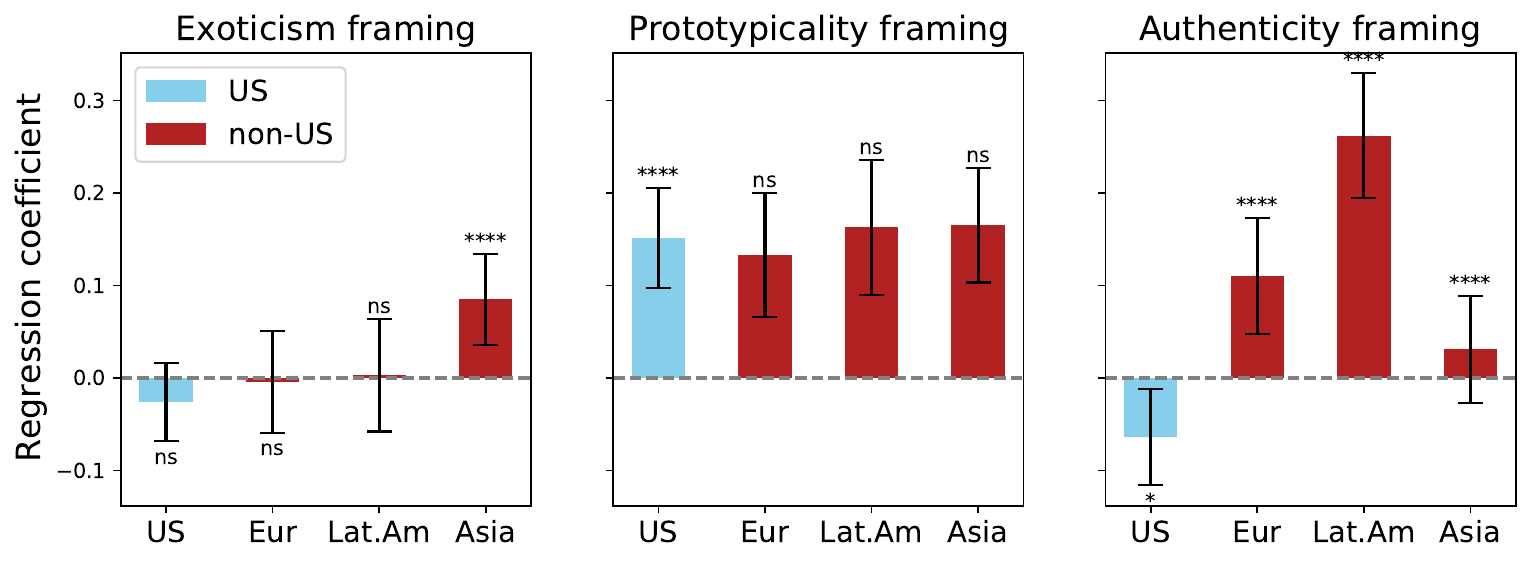}
    \caption{Coefficients from linear mixed effects models estimating othering from cuisine type, with random effects per user, showing that \textsc{As} is framed as more exotic and immigrant cuisines are framed as more authentic.}
    \label{fig:user_cont_study1_results}
\end{figure}

\begin{figure}[h!]
    \centering
    \includegraphics[scale=0.27]{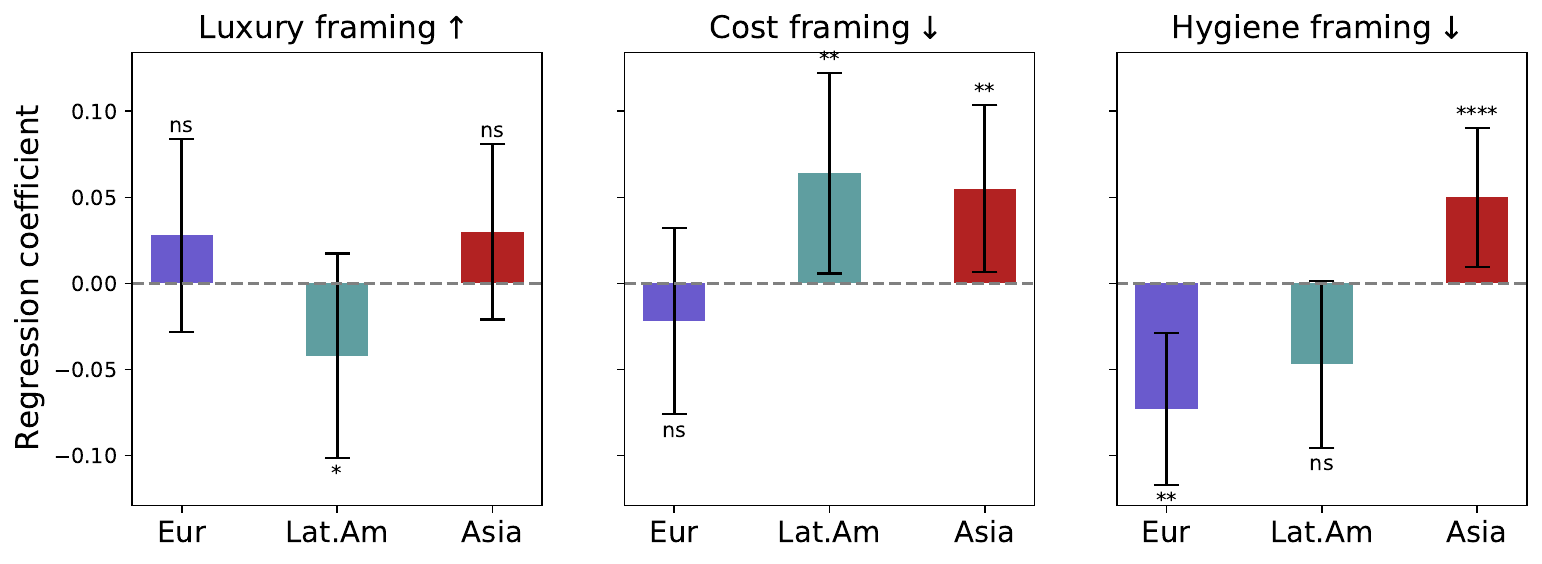}
    \caption{Coefficients from linear mixed effects models estimating high $\uparrow$ vs. low status $\downarrow$ framing from cuisine type, with random effects per user.}
    \label{fig:user_cont_study2_results}
\end{figure}

\xhdr{User-controlled analysis}\label{ssec:user_controlled} We replicate analyses on 17K reviews of 89 high-volume users contributing $\geq$100 reviews and $\geq$10 per region. We fit linear mixed effects models with random user effects and fixed cuisine effects. Results, when significant, corroborate main results (Figures \ref{fig:user_cont_study1_results}-\ref{fig:user_cont_study2_results}).

% \subsection{Top-cuisine removed analysis}\label{ssec:top_cuisine_removed}

%\input{formatting_instructions.tex}

% \subsection{Full LM analysis}

% We perform analyses on synthetic reviews balanced for sentiment (actual consumer reviews skew highly positive) and find similar results to Study 3, with additional effects of negative and positive hygiene framing in reviews of \textsc{As} restaurants.

% \begin{table}[]
%     \centering
%     \begin{tabular}{c|c}
%          &  \\
%          & 
%     \end{tabular}
%     \caption{\todo{[Yiwei] add updated LM results once additional data is collected}}
%     \label{tab:my_label}
% \end{table}
\newpage

\end{document}